\documentclass[10pt,twocolumn,letterpaper]{article}

\def\maketitlesupplementary
   {
   \newpage
       \twocolumn[
        \centering
        \Large
        \textbf{Adversarial Attention Perturbations for Large Object Detection Transformers}\\
        \vspace{0.5em}Supplementary Material \\
        \vspace{1.0em}
       ] 
   }

\usepackage{iccv}
\usepackage{times}
\usepackage{epsfig}
\usepackage{graphicx}
\usepackage{amsmath}
\usepackage{amssymb}
\usepackage{booktabs}
\usepackage{multirow}
\usepackage{soul}
\usepackage{algpseudocode}
\usepackage{algorithm}
\usepackage{blindtext}
\usepackage{chngcntr}
\counterwithin*{section}{part}

\usepackage[accsupp]{axessibility}  

\usepackage[table, x11names]{xcolor}

\usepackage[pagebackref=true,breaklinks=true,letterpaper=true,colorlinks,bookmarks=false]{hyperref}

\iccvfinalcopy 

\def\iccvPaperID{14244} 

\ificcvfinal\fi

\begin{document}

\title{Adversarial Attention Perturbations for Large Object Detection Transformers}

\author{Zachary Yahn\textsuperscript{1}, Selim Furkan Tekin\textsuperscript{1}, Fatih Ilhan\textsuperscript{1}, Sihao Hu\textsuperscript{1}, \\ Tiansheng Huang\textsuperscript{1}, Yichang Xu\textsuperscript{1},  Margaret Loper\textsuperscript{2}, Ling Liu\textsuperscript{1}  \\
\textsuperscript{1}Georgia Institute of Technology, Atlanta, GA\\
\textsuperscript{2}Georgia Tech Research Institute, Atlanta, USA\\
{\tt\small \{zachary.yahn, stekin6, filhan, sihaohu, thuang, xuyichang\}@gatech.edu}\\
{\tt\small margaret.loper@gtri.gatech.edu, ling.liu@cc.gatech.edu}}


\maketitle
\ificcvfinal\thispagestyle{empty}\fi

\begin{abstract}
\noindent

Adversarial perturbations are useful tools for exposing vulnerabilities in neural networks. Existing adversarial perturbation methods for object detection are either limited to attacking CNN-based detectors or weak against transformer-based detectors. This paper presents an Attention-Focused Offensive Gradient (AFOG) attack against object detection transformers. By design, AFOG is neural-architecture agnostic and effective for attacking both large transformer-based object detectors and conventional CNN-based detectors with a unified adversarial attention framework.  This paper makes three original contributions. First, AFOG utilizes a learnable attention mechanism that focuses perturbations on vulnerable image regions in multi-box detection tasks, increasing performance over non-attention baselines by up to 30.6\%. Second, AFOG's attack loss is formulated by integrating two types of feature loss through learnable attention updates with iterative injection of adversarial perturbations. Finally, AFOG is an efficient and stealthy adversarial perturbation method. It probes the weak spots of detection transformers by adding strategically generated and visually imperceptible perturbations which can cause well-trained object detection models to fail. Extensive experiments conducted with twelve large detection transformers on COCO demonstrate the efficacy of AFOG. Our empirical results also show that AFOG outperforms existing attacks on transformer-based and CNN-based object detectors by up to 83\% with superior speed and imperceptibility. Code is available at: \href{https://github.com/zacharyyahn/AFOG}{Link}.
\end{abstract}

\vspace{-10pt}
\section{Introduction}
\begin{figure}[h]
    \centering
    \includegraphics[width=\columnwidth]{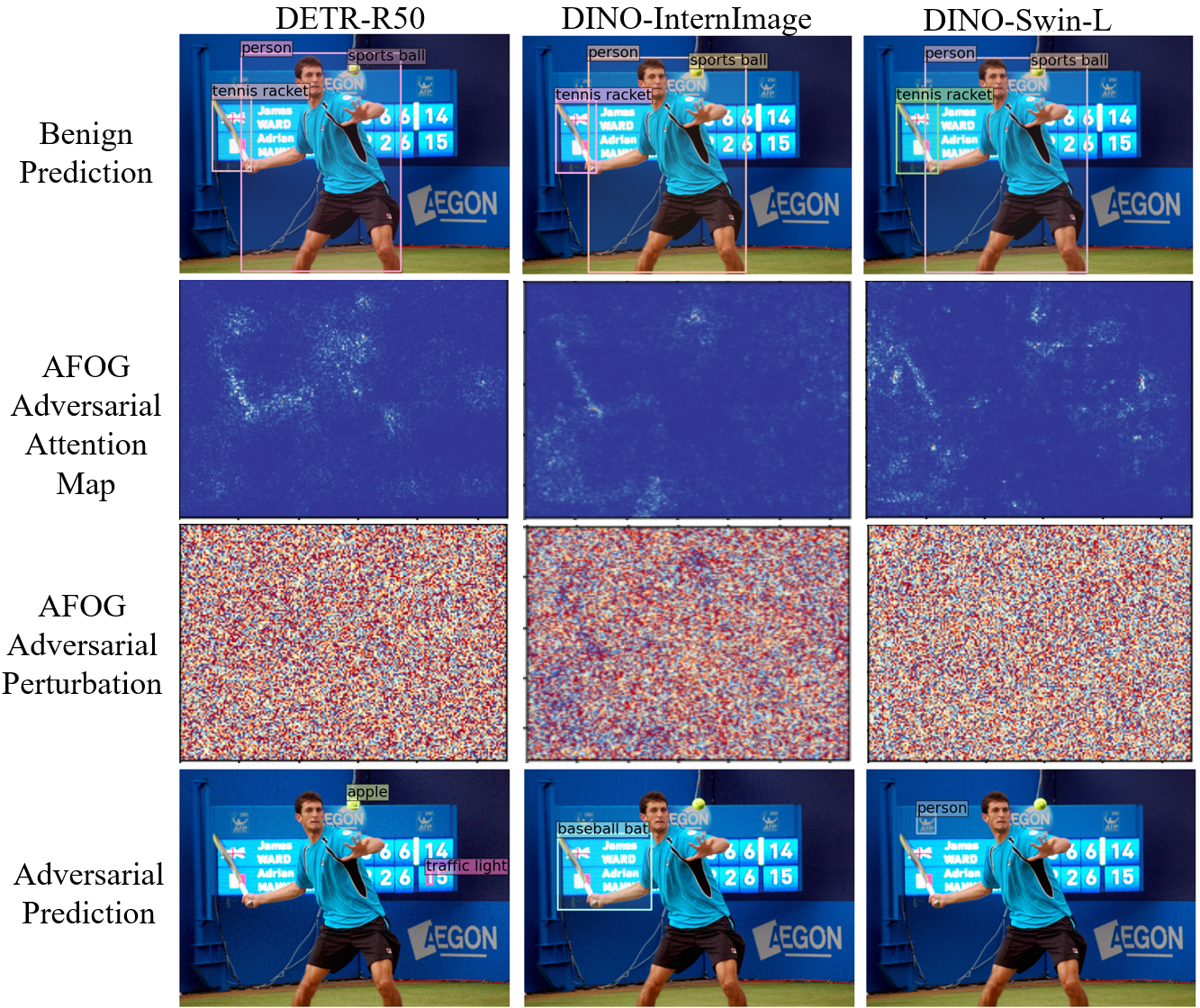}
    \caption{\small Successful AFOG attacks against three object detection transformers. White pixels in the attention map indicate learned areas of high importance, often corresponding to objects in the input image. This example illustrates that AFOG's adversarial attention learns which regions of an image are most susceptible to adversarial perturbations. Even for the same input, learnable adversarial attention maps may differ under different models.} 
    \label{fig:demo}
    \vspace{-15pt}
\end{figure}

Transformer-based neural architectures and algorithms have blossomed in recent years, enhancing computer vision tasks including object detection. Attention is the core of the transformer architecture~\cite{vaswani2017transformer}. Attention allows a detector to focus on specific regions of images based on objects of interest, effectively predicting the presence and location of each potential object. Powered by various attention mechanisms, detection transformers 
can handle overlapping objects more effectively than some older methods because they can capture long-range dependencies in an image, allowing models to reason about the relationships between objects. Since modern object detection transformers such as Swin \cite{liu2021Swin} and DETR \cite{carion2020} significantly outperform traditional convolutional neural network (CNN)-based models~\cite{cocoleaderboard}, represented by Faster R-CNN~\cite{ren2016frcnn}, SSD~\cite{wei2016ssd} and YOLO-v3~\cite{redmon2018yolov3}, there is a pressing need to investigate and understand the vulnerabilities of large detection transformers in the presence of adversarial perturbations.
Adversarial perturbations are helpful instrumentation for 
exposing vulnerabilities of large detection transformers, making them useful mechanisms for motivating more robust transformer models for object detection. 

Existing attack methods struggle to disrupt transformer-based object detectors. Surrogate-based attacks (also known as black-box attacks), such as UEA~\cite{wei2019uea} and RAD~\cite{RAD}
generate perturbations on surrogate models and then test the adversarial effect against victim detectors that have similar detection architecture at inference. However, surrogate-based attacks suffer poor attack performance when the victim architectures are not similar to the surrogate models used to generate adversarial perturbations. 
Victim-based (also known as white-box) attacks, such as EBAD~\cite{cai_ebad} and OATB~\cite{leng_oatb}, generate adversarial attacks by performing direct inference against victim models.
Recent studies, such as AttentionFool \cite{lovisotto_attnfool}, 
specialize in transformer-based victim models, but these transformer-only attacks are not applicable to convolutional detectors such as YOLO \cite{redmon2018yolov3}.


In this paper, we present an Attention-Focused Offensive Gradient (AFOG) attack, which is victim architecture agnostic and effective in attacking both advanced object detection transformers and traditional CNN-based detectors. 
AFOG by design offers three novel characteristics. First, inspired by a transformer's self-attention, we utilize a learnable attention mechanism to enable AFOG to adaptively focus its adversarial perturbations on vulnerable image areas (those areas where perturbations have the greatest effect on the output) in multi-box detection tasks. Second, we formulate the attack loss function of AFOG by integrating learnable feature-map-based attention updates with iterative injection of adversarial perturbations. Finally, we design AFOG to be an efficient and yet stealthy adversarial perturbation method. By efficient, we want AFOG to rapidly generate adversarial perturbations in minimal iterations. By stealthy, we want AFOG to generate small amounts of perturbations that are visually imperceptible and yet can cause well-trained object detectors to produce erroneous detection results. 
Figure~\ref{fig:demo} shows an example of successful AFOG attacks against three detection transformers. Row 1 shows benign detection from three state-of-the-art detectors, Row 2 and Row 3 show AFOG adversarial perturbations on the three detectors and their respective AFOG adversarial attention maps, and Row 4 shows the detection results under our AFOG attack. 
We validate AFOG with extensive experiments on twelve state-of-the-art object detection transformers with the COCO benchmark~\cite{coco} and three popular families of traditional CNN-based object detectors. 
The results show that AFOG is consistently effective across all twelve detection transformers compared to existing methods, achieving high attack success rate and reducing benign mAP (mean Average Precision) by up to 37.79$\times$. 

\section{Related Work}
Existing adversarial perturbations against object detectors can be broadly categorized into surrogate-model based (also known as black-box) and victim-model based (white-box) methods. Surrogate-based approaches in literature
tend to rely on the assumption that the surrogate model is similar to the victim model. RAD \cite{chen_rad}, GHFD \cite{wang_ghfd}, and UEA \cite{wei2019uea} generate adversarial examples on a surrogate FRCNN model, then attack other CNN-based detectors such as YOLO and SSD. However, 
these attacks show poor performance against object detection transformers~\cite{wang_ghfd}. For victim-model based methods,  
GARSDC \cite{liang_garsdc} 
is more effective but can require more than 3000 iterations to converge. GALD \cite{li_gald} first attacks vision transformer classifiers and then transfers the classification-based adversarial perturbations to object detectors of similar transformer architecture, and hence is less effective by comparison. 
We classify EBAD \cite{cai_ebad} as a victim-based attack because it uses an ensemble of surrogate models to attack a victim, but requires access to the victim's loss function for its nested ensemble optimization. EBAD shows poor performance when transferring perturbations from a DETR surrogate to a DETR victim \cite{nguyen_survey_2024}.
AttentionFool \cite{lovisotto_attnfool} is a recent victim-based attack targeting dot-product self-attention
against DETR~\cite{li2022vitdet} with a ResNet-101 backbone. However, AttentionFool shows inconsistent performance against DETR with a ResNet-50 backbone \cite{lovisotto_attnfool}. AttentionFool is also not applicable 
to convolutional models such as YOLO \cite{redmon2018yolov3} because it exclusively targets self-attention mechanisms. TOG \cite{chow2020tog} targets a prediction's objectness score and introduces vanishing and fabrication attack modes. OATB \cite{leng_oatb} uses a ``division map" that statically emphasizes image regions during perturbation based on object location priors. DBA \cite{lian_dba} prioritizes perturbations against image backgrounds to enhance imperceptibility, though it shows almost no effect against a Swin transformer for object detection~\cite{liu2021Swin}. 
In comparison, AFOG adversarial perturbations show strong performance against a wide variety of transformer- and CNN-based detector victims while maintaining superior imperceptibility and efficiency.

\section{Methodology}
\subsection{Problem Definition}
\noindent
Given a victim detector $f_D(\vartheta, x)$, where $x\in \mathcal{D}$ is a victim image $x$ and $\mathcal{D}$ is the test set, let $x$ contain $N_x$ objects to be detected, denoted by $\mathcal{O}_x=\{O_1, O_2, \dots,{O_N}_x\}$. Each object $O_i$ ($1\leq i\leq N_x$) is a recognition target for the detector $f_D(\vartheta, x)$. Under benign scenarios, let $(B_i, C_i)$ denote a ground truth object $O_i$ with bounding box $B_i$ and class label $C_i$. Let $K$ denote the total number of ground truth classes, and $C_i \in \{1,2,\dots,K\}$, e.g., $K=21$ for the VOC dataset~\cite{pascalvoc}, including the background class. Given input image $x$,
$f_D(\vartheta, x)$ outputs the benign prediction of $N$ detected objects, denoted by $\mathcal{R}[f_D(\vartheta, x)]=\{(b_i, c_i) | i=1,\dots,N_x\}$. Each detected object $o_i$ is associated with its predicted bounding box $b_i$ and predicted class label $c_i$. $(b_i, c_i)$ is evaluated as a correct prediction if the intersection over union (IoU) of $B_i$ and $b_i$ is greater than the detection threshold $\gamma$ (usually set to $0.5$), and $C_i = c_i$ ($1\leq i\leq N_x$, $x\in \mathcal{D}$). The overall detection accuracy is measured using mAP (mean Average Precision) on the entire test dataset $\mathcal{D}$. 

Let $x_{adv}$ denote the adversarial example generated by injecting a sequence of adversarial perturbations to $x$ through an iterative attention-based learning mechanism. The goal of our AFOG attack on detector $f_D(\vartheta, x)$ is to find $x_{adv}$ that maximizes the success rate of falsifying the predictions of all object recognitions for all images in $\mathcal{D}$, i.e. \\
\vspace{2pt}
\indent ${argmax}_{x\in \mathcal{D}, i\in N_x}, \{ (b_i, c_i) \in \mathcal{R}[f_D(\vartheta, x_{adv})] \\
\indent \indent 
(IOU(B_i, b_i) < \gamma \lor C_i \neq c_i), min||x-x_{adv}||_p \}$. 
\\
The formula indicates that for each detected object $o_i$, denoted by ($b_i, c_i$) in victim image $x$, the AFOG attack is successful on $o_i$ if the IoU of $B_i$ and $b_i$ is less than the detection threshold $\gamma$ (usually set to $0.5$) or $c_i \in \{1,2,\dots,K\}(C_i\neq c_i)$, and $x_{adv}$ also satisfies the distortion constraint of $min||x-x_{adv}||_p$, where $p$ is typically defined by $L_2$ norm, $L_0$ norm, or $L_\infty$ norm. 

\begin{figure}[h]
    \centering
    \includegraphics[width=\columnwidth]{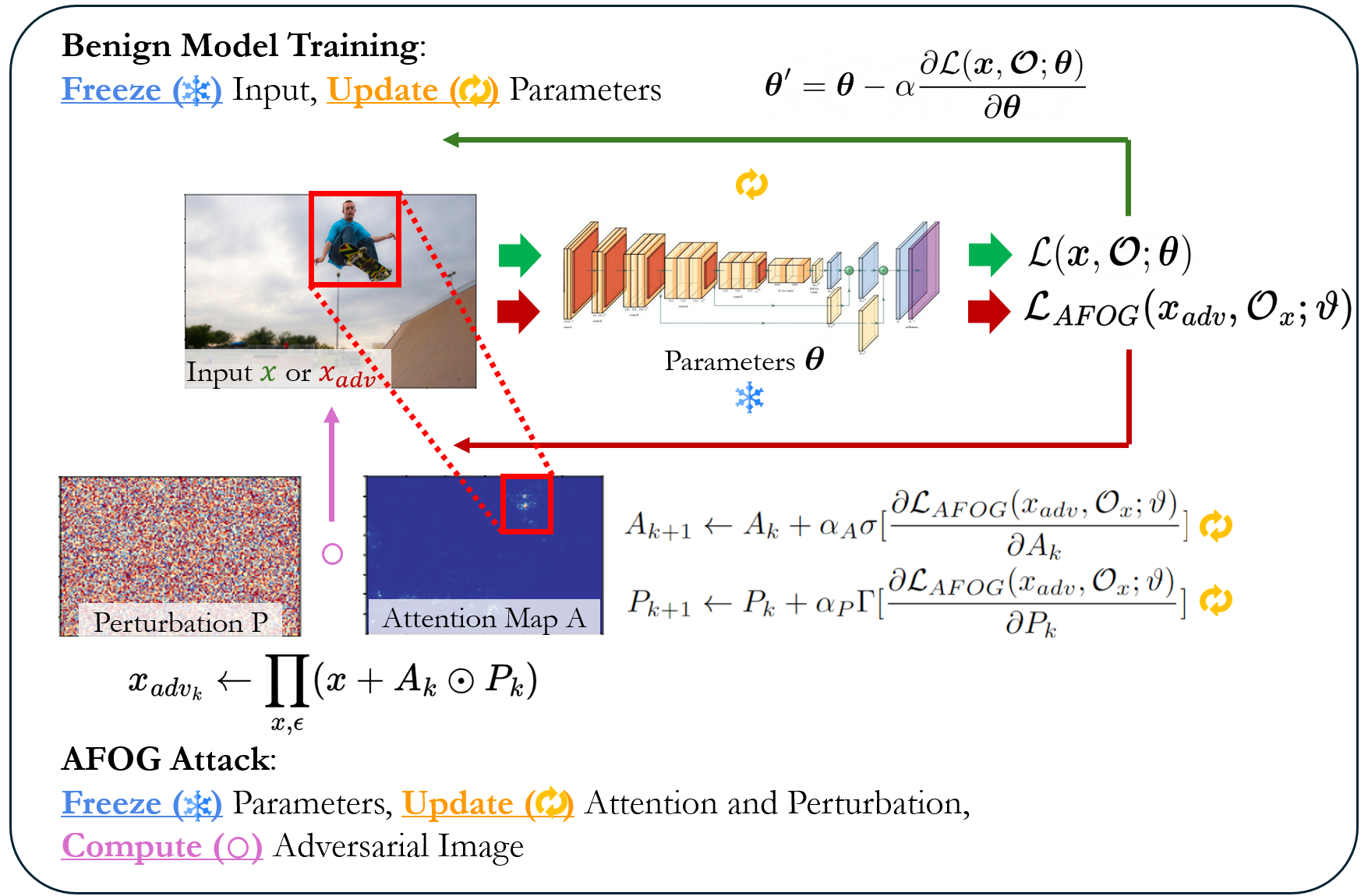}
    \caption{AFOG attack framework. During attack iteration the adversarial perturbation is updated via bounding-box and class loss gradient propagation. An adversarial attention map concurrently learns to apply per-pixel amplification or dampening to the perturbation, focusing its effects on vulnerable image regions.}
    \label{fig:framework}
    \vspace{-10pt}
\end{figure}

\subsection{Adversarial Loss Optimization}
\noindent
AFOG attacks the victim model via an iterative projected gradient descent method \cite{madry2018pgd}. Figure \ref{fig:framework} provides an illustration of the AFOG attack framework. At the initialization step, the attack first propagates the unaltered victim image $x$ through the feed-forward network (FFN) of the victim model $f_D(\vartheta, x)$ to obtain its benign prediction $\mathcal{O}_x = \{(b_i, c_i) | i=1,\dots,N_x\}$. 

In a scenario where no ground truth is available, the attack assumes these benign predictions to be correct labels for the input image $x$. For the remaining section, we will use $(B_i, C_i)$ in the context of performing an attack on $f_D(\vartheta, x)$ to obtain $x_{adv}$. Upon initialization for each input image $x\in \mathcal{D}$, the AFOG attack corrupts the victim image $x$ by adding perturbations generated via an element-wise multiplication of two components using Equation~\ref{eq:calc_perturbation}: 
\vspace{-6pt}
\begin{equation}
    \label{eq:calc_perturbation}
    x_{adv_{k}} \gets \prod\limits_{x, \epsilon}(x + A_{k} \odot P_{k})
    \vspace{-6pt}
\end{equation}
\noindent 
$A$ is the attention map, and $P$ is the perturbation map. Here $\odot$ represents the Hadamard matrix product and $\Pi$ is a projection onto a hypersphere centered on unaltered victim image $x$ with the radius $\epsilon$ as the maximum perturbation budget. $A$ and $P$ are initialized according to Equation~\ref{eq:init}.
\vspace{-6pt}
\begin{equation}
    \label{eq:init}
    A_0 \sim \mathop{1},    P_0 \sim Random(-\epsilon, \epsilon)
\vspace{-6pt}
\end{equation}
\noindent 
$Random(-\epsilon, \epsilon)$ is a uniform random distribution. The perturbed image $x_{adv}$ propagates through the FFN of the victim model. The attack loss function $\mathcal{L}_{AFOG}$ assesses the adversarial output $\mathcal{O}_{x_{adv}}$ and benign output $\mathcal{O}_x$, computing a loss that reflects the attack's progress in corrupting image $x$. This attack loss is formulated in Equations~\ref{eq:AFOG-loss}, \ref{eq:bbox-loss}, and \ref{eq:cls-loss}.
\begin{equation}
    \label{eq:AFOG-loss}
    \begin{split}
    \mathcal{L}_{AFOG}(x_{adv}, \mathcal{O}_x; \vartheta) &  
   = \mathcal{L}_{bbox}(x_{adv}, \mathcal{O}_x; \vartheta) \\
   & + \mathcal{L}_{cls}(x_{adv}, \mathcal{O}_x; \vartheta)
    \end{split}
\end{equation}
\begin{equation}
  \label{eq:bbox-loss}
    \mathcal{L}_{bbox}(x_{adv}, \mathcal{O}_x; \vartheta)
   = \sum_{i=1}^{N_x} [f_{\vartheta}(x, o_i) - f_{\vartheta}(x_{adv}, o_{adv_i})]  
\end{equation}
\begin{equation}
  \label{eq:cls-loss}
    \mathcal{L}_{cls}(x_{adv}, \mathcal{O}_x; \vartheta) \\ 
   = \sum_{i=1}^{N_x} [f_{\vartheta}(x, c_i) - f_{\vartheta}(x_{adv}, c_{adv_i})] 
\end{equation}

\begin{figure*}
    \centering
    \includegraphics[width=1.4\columnwidth]{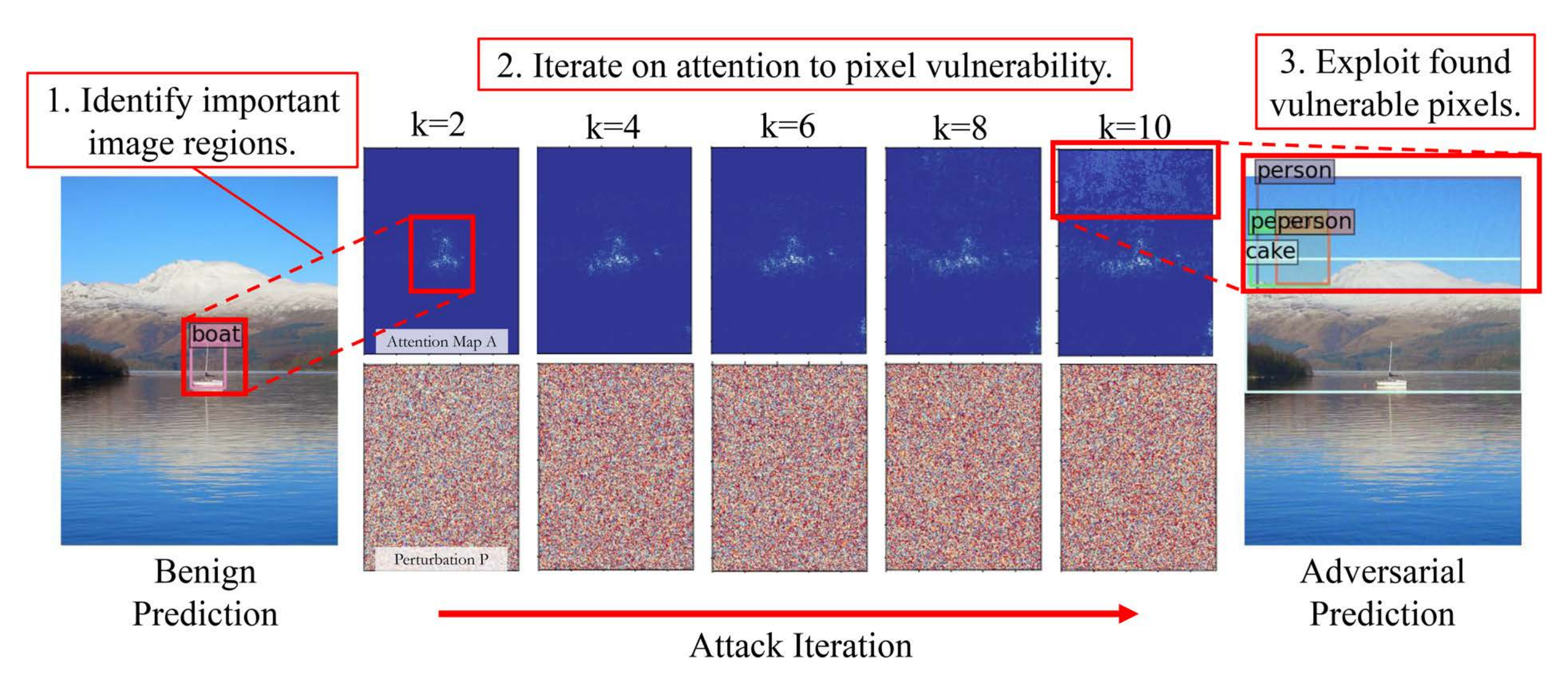}
    \caption{Example attention map generation on DETR-R101 for k=10 iterations. AFOG attention initially learns important regions of the image, then iteratively updates to achieve an effective attack. It finds vulnerable pixels in foreground objects as well as background regions, including in unexpected or unintuitive areas such as the sky above the boat.}
    \label{fig:attention}
\end{figure*}

Given a victim image $x$, the optimized attack loss $\mathcal{L}_{AFOG}$ is attained by making the model incorrectly predict every target object. We achieve this loss optimization by both falsifying every target object's bounding box prediction and its class label prediction. This can be represented by optimizing an adversarial bounding box loss $\mathcal{L}_{bbox}$ and an adversarial class label prediction loss $\mathcal{L}_{cls}$ for all $N_x$ target objects. Conceptually, this suppresses the confidence of the original correct bounding box and class label prediction, while increasing the confidence of the adversarial prediction of an incorrect bounding box or class label. 

Recall Figure~\ref{fig:framework}: the AFOG attack freezes the model parameters $\vartheta$ and uses the gradient of the AFOG attack loss to update the attention map $A$ and the perturbation $P$ by backpropagation, generating the next iteration of the adversarial perturbation. A newly corrupted image $x_{adv}$ is created upon injecting the updated adversarial perturbation by following Equation \ref{eq:calc_perturbation}, where A and P are updated via Equations~\ref{eq:backprop_a} and \ref{eq:backprop_p}:
\vspace{-6pt}
\begin{equation}
    \label{eq:backprop_a}
    A_{k+1} \gets A_k + \alpha_{A}\sigma[\frac{\partial\mathcal{L}_{AFOG}(x_{adv}, \mathcal{O}_x; \vartheta)}{\partial A_k}]
\end{equation}
\vspace{-8pt}
\begin{equation}
    \label{eq:backprop_p}
    P_{k+1} \gets P_k + \alpha_{P}\Gamma[\frac{\partial\mathcal{L}_{AFOG}(x_{adv}, \mathcal{O}_x; \vartheta)}{\partial P_k}]
\end{equation}
\noindent
$\alpha_A$ and $\alpha_P$ denote the attention map learning rate and the perturbation learning rate respectively, $\sigma$ is a normalization function, $\Gamma$ is a sign function, $\mathcal{L}_{AFOG}$ is the attack loss, $\vartheta$ is the model parameters, and $\mathcal{O}_x$ is the model's benign predictions from input $x$. 
This attack process repeats until the number of attack iterations is reached. Algorithm \ref{alg:afog} provides the pseudocode. 

\begin{algorithm}
\caption{AFOG attack on an input image.}\label{alg:afog}
\begin{algorithmic}[1]
\Require 
Victim image $x\in \mathcal{D}$, test-set $\mathcal{D}$, 
Victim pre-trained model $f_D(\vartheta)$,
Perturbation step size $\alpha_P$, 
Attention step size $\alpha_A$, 
Number of iterations $T$, 
Maximum perturbation $\epsilon$.
\State Initialize $\mathcal{O}_x \leftarrow f_D(x; \vartheta)$
\State Initialize attention map $A_0\leftarrow 1$;
\State Initialize perturbation $P_0\leftarrow Random(-\epsilon, \epsilon)$;
\State Initialize step variable $k \gets 1$;
\While{$k \leq T$}
    \State Attack image ${x_{adv}}_{k} \gets \prod\limits_{x, \epsilon}(x + A_{k} \odot P_{k}$);
    \State Forward propagate $x_{adv}$ through $f_{D}(\vartheta, x_{adv})$;
    \State Compute bbox-loss $ \mathcal{L}_{bbox}(x_{adv}$, $\mathcal{O}_x; \vartheta)$;
    \State Compute cls-loss $\mathcal{L}_{cls}(x_{adv}$, $\mathcal{O}_x; \vartheta)$; 
\State $\mathcal{L}_{AFOG}(x_{adv}, \mathcal{O}_x; \vartheta)=$ bbox-loss + cls-loss;
    \State Calculate losses with respect to $A_k$ and $P_k$: $\mathcal{L}_A(x_{adv}, \mathcal{O}_x; \vartheta), \mathcal{L}_P(x_{adv}, \mathcal{O}_x; \vartheta)$;
    \State Normalize attention loss $\mathcal{L}_A \gets$ Norm$(\mathcal{L}_A)$;
    \State Take sign of perturbation loss $\mathcal{L}_P \gets$ Sign$(\mathcal{L}_P)$;
    \State $A_{k+1} \gets A_k - \alpha_A\mathcal{L}_A$;
    \State $P_{k+1} \gets P_k - \alpha_P\mathcal{L}_P$;
    \State $k \gets k+1$;
\EndWhile
\State ${x_{adv}}_{k+1} \gets \prod\limits_{x, \epsilon}(x + A_{k+1} \odot P_{k+1})$;
\State \Return $x_{adv}$
\end{algorithmic}
\end{algorithm}

\subsection{Adversarial Attention Mechanism}
\vspace{-3pt}
\noindent 
A key innovation of our AFOG method is empowering perturbation generation with a learnable attention mechanism. Motivated by the intuition that certain parts of a victim image are more susceptible to adversarial perturbation than others, we add an attention map to focus the perturbation on vulnerable pixels. The AFOG attack concurrently learns the adversarial attention map alongside the perturbation to iteratively maximize the attack loss $\mathcal{L}_{AFOG}$ (recall Equation~3). Figure \ref{fig:attention} gives an illustrative example of the iterative learning of AFOG's adversarial attention map and the corresponding perturbation. Unlike focusing mechanisms in other methods such as \cite{wei2019uea}, AFOG's attention map dynamically updates during attack iteration, leaving it unconstrained by assumptions about foreground \cite{leng_oatb}, background \cite{lian_dba}, or region proposal \cite{wei2019uea} importance from static methods. We posit that pixel importance may not correspond to human intuition, and design AFOG to iteratively learn pixel importance instead. We observe that in early iterations the attention mechanism tends to focus on the primary objects in an image, and then branches out to affect surrounding areas as the attack progresses. This adaptability is a key feature of learnable attention, showcasing an advantage over static attention maps. 

\vspace{-3pt}
\subsection{Special Cases of the AFOG Attack}
\vspace{-3pt}
\noindent 
We explore two special cases of our AFOG attack, each of which targets a specific malicious detection behavior in the victim detector. The first special case is an object vanishing attack, coined AFOG-V, which attempts to attack the objectness detection task of multi-box object detection. The goal of this special case AFOG attack is to make the victim model unable to detect any object, making all object detections vanish for victim image $x$. We implement the AFOG-V attack by changing the initialization: we use an empty set instead of forward propagation of $x$ to obtain the benign detection results for the $N_x$ objects as the assumed ground-truth in AFOG-V attack. Let $\varnothing$ denote an altered version of $\mathcal{O}_x$ that contains no predictions. The formulation of the AFOG-V attack loss function is given by Equation \ref{eq:afogv}:
\vspace{-8pt}
\begin{equation}
    \label{eq:afogv}
    \begin{split}
    \mathcal{L}_{AFOG_V}(x_{adv}, \mathcal{O}_x; \vartheta) 
    = -\mathcal{L}_{bbox}(x_{adv}, \varnothing; \vartheta)
    \\
    - \mathcal{L}_{cls}(x_{adv}, \varnothing; \vartheta)
    \end{split}
  \vspace{-8pt}
\end{equation}
\noindent 
The second special case is object fabrication, coined as AFOG-F, which attempts to attack the bounding-box detection task by generating perturbations that lead to spurious detections (i.e.,  false positives). Similarly, we revise the AFOG attack process: instead of keeping the benign predictions with the confidence score above a certain tunable threshold (0.5 by default), for AFOG-F, we remove this threshold and allow the benign detection set $\mathcal{O}_x$ to include a much larger set of ``ground truth'' targets, which inevitably include erroneous detections. Accordingly the AFOG-F loss function is given in Equation \ref{eq:afogf}:
\vspace{-8pt}
\begin{equation}
    \label{eq:afogf}
    \begin{split}
    \mathcal{L}_{AFOG_F}(x_{adv}, \mathcal{O}_x; \vartheta) 
    = -\mathcal{L}_{bbox}(x_{adv}, \mathcal{O}_F; \vartheta)
    \\
    - \mathcal{L}_{cls}(x_{adv}, \mathcal{O}_F; \vartheta)
    \end{split}
    \vspace{-8pt}
\end{equation}
\noindent 
where $\mathcal{O}_F$ is a modified version such that the likelihood score of each prediction is set to 1.0. We explore these two specialized cases of our AFOG attack to gain a deeper understanding of the adverse effects of adversarial perturbations on different detection transformers.  

\begin{figure*}[h]
    \centering
    \includegraphics[width=2.1\columnwidth]{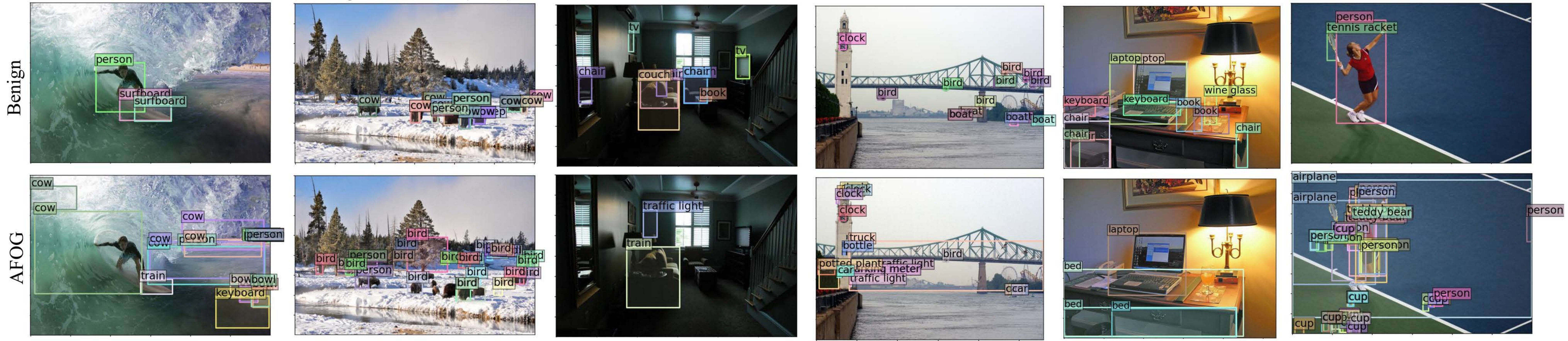}
     \vspace{-4pt}
    \caption{Visualization with six examples, comparing benign scenario and AFOG attack results on Detection Transformer (DETR) with ResNet-50 backbone. Six diverse test images contain objects with varying classes, scales, perspectives, lighting, motion, and textures.}
    \label{fig:examples}
    \vspace{-8pt}
\end{figure*}

\section{Experiments}
\subsection{Experimental Setup}
\noindent 
We use the Common Objects in Context (COCO) 2017 \cite{coco} test-dev split to evaluate the performance of our attack on contemporary object detectors. COCO 2017 is a standard benchmark for evaluating object detector performance, and its test-dev set has 80 object categories across 5,000 images. We use the PASCAL Visual Objects Challenge (VOC) 2007 \cite{pascalvoc} dataset to compare our attack with the state-of-the-art attacks on CNN-based object detectors, because existing attacks designed only for CNN-based detectors all report their evaluation with VOC. The VOC devkit split contains 4,952 images with 20 object classes. All experiments were performed on an NVIDIA A100. We list our models and their benign performance in Table~\ref{tab:main_results}. We provide model details in Section \ref{sec:models} of the Supplementary Material. 


We select a variety of model sizes, ranging from Detection Transformer (DETR) \cite{carion2020}, a lightweight 40 million parameter model, to EVA \cite{EVA}, a vision-focused foundation model with over one billion parameters. We use the Detrex framework \cite{ren2023detrex}, built on top of Detectron2 \cite{wu2019detectron2}, to standardize our model implementations via DINO \cite{zhang2022dino}. DINO and Detrex adapt many general purpose vision models to object detection. The Detrex framework standardizes all images in $[0, 1]$. We also attack the original versions of several transformer models to demonstrate that the AFOG attack also works beyond the Detrex framework. All 12 transformer models and FRCNN use PyTorch \cite{pytorch} implementations, whereas SSD300 \cite{wei2016ssd} and YOLOv3 \cite{redmon2018yolov3} use Tensorflow \cite{tensorflow2015-whitepaper}. We use mean average precision (mAP) to evaluate both benign and attacked performance. The Average Precision (AP) is given by interpolating the product of precision and recall at several decision thresholds. The mean AP (mAP) is the average of APs over the number of object classes. A lower mAP indicates that the detector is less effective for the object detection task. The victim detector's mAP under adversarial perturbations reflects the degree of degradation in detection performance for the victim detector when compared with its benign mAP. We measure imperceptibility by reporting the average distortion of the attack via four metrics: $L_2$ norm, $L_0$ norm, semantic structural similarity index (SSIM), and mean distortion $\mu_\Delta$. We compute each metric per image and report the average over all images in COCO 2017 test-dev. Our experiments in the next section show that setting the total number of attack iterations to 10 is effective for all 12 state-of-the-art detection models on the COCO benchmark.  

\subsection{Effectiveness Comparison of AFOG Attack}
\noindent
Table~\ref{tab:main_results} shows the effectiveness of the AFOG attack and its two special extensions AFOG-V and AFOG-F on each of the twelve detection transformers with varying model sizes (Col. 2). We use their benign mAP scores (Col. 3) as the reference for the evaluation. We make three observations: (i) under AFOG attacks all 12 transformers suffer a drastic performance drop in mAP scores, ranging from $2.5\times$ reduction to $37.8\times$ reduction. (ii) Compared to the AFOG generic attack, the special extension AFOG-V offers stronger attack success rates in terms of mAP reduction for 11 out of 12 transformers, except DETR-R50~\cite{carion2020}; and the special extension AFOG-F offers stronger attack success rates for 7 out of 12 transformers, except DETR-R50~\cite{carion2020}, DETR-R101~\cite{carion2020}, Deformable-DETR~\cite{zhu2020deformable}, R50 (DINO)~\cite{he2016resnet}, Swin-L (DINO)~\cite{liu2021Swin} (iii) For AlignDETR ~\cite{cai2023aligndetr}, AFOG-F achieves a mAP of 1.36 and outperforms both AFOG (mAP of 18.13) and AFOG-V (mAP of 1.64), and all three attacks significantly reduce the benign mAP by a factor ranging from $2.8\times$ to $37.8\times$.

We compare AFOG against eleven benchmark attacks on DETR and Swin in Table \ref{tab:related_comparisons}. We observe that AFOG achieves superior performance on both DETR-R50 and Swin-L, notably recording an 82.7\% improvement over the next strongest attack on Swin. AFOG also uses the smallest perturbation budget and fewest iterations, indicating its superior imperceptibility and speed.

\bgroup
\addtolength{\tabcolsep}{-0.1em}
\begin{table}[h]
  \fontsize{7pt}{7pt}\selectfont
  \centering
  \caption{AFOG effectiveness over 12 detection transformers, measured by mAP on perturbed images. (*) indicates DINO framework used with the corresponding backbone for object detection.}
  \begin{tabular}{cccccc}
    \toprule
    Model & Params (M) & Benign & AFOG & AFOG-V & AFOG-F \\
    \midrule
    DETR-R50 \cite{carion2020} & 39.8 & 42.1 & 4.1 & 4.5 & 9.8 \\
    DETR-R101 \cite{carion2020} & 76.0 & 43.5 & 5.2 & 5.1 & 11.3 \\
    Deform.-DETR \cite{zhu2020deformable} & 40.0 & 44.5 &  4.8 & 1.5 & 7.1 \\
    R50* \cite{he2016resnet} & 47.6 & 49.2 &  5.3 & 1.5 & 6.3 \\
    AlignDETR \cite{cai2023aligndetr} & 47.6 & 51.4 & 18.1 & 1.6 & 1.4\\
    ViTDet* \cite{li2022vitdet} & 108.1 & 54.9 & 3.8 & 0.9 & 2.8 \\
    ConvNeXt* \cite{liu2022convnext} & 219.0 & 55.4 & 3.9 & 1.9 & 3.1\\
    Swin-L* \cite{liu2021Swin} & 217.2 & 56.8 & 7.3 & 2.4 & 8.6\\
    InternImage* \cite{wang2022internimage} & 241.0 & 56.9 & 7.3 & 2.8 & 5.1 \\
    FocalNet* \cite{yang2022focal} & 228.9 & 58.5 & 7.3 & 2.5 & 5.1\\
    EVA* \cite{EVA} & 1037.2 & 62.1 & 12.2 & 4.1 & 8.7\\
    DETA \cite{ouyangzhang2022nms} & 218.8 & 62.9 & 25.6 & 3.7 & 4.3 \\
    \bottomrule
  \end{tabular}
  \label{tab:main_results}
\end{table}
\egroup 

\bgroup
\addtolength{\tabcolsep}{-0.1em}
\begin{table}[h]
  \def\arraystretch{1.2}
  \fontsize{7pt}{7pt}\selectfont
  \centering
  \caption{Comparison benchmark of AFOG against other state-of-the-art object detection attacks on DETR and Swin. Results are theirs. (*) indicates results from \cite{wang_ghfd}. ($\dagger$) indicates results from \cite{nguyen_survey_2024}. (-) indicates no result.}
  \begin{tabular}{cccccc}
    \toprule
     & & & & \multicolumn{2}{c}{Adversarial mAP} \\
    \cline{5-6} Attack & Type & Pert. Budget & Iters. & \rule[1pt]{0pt}{7pt} DETR-R50 & Swin \\
    \midrule
    GARSDC \cite{liang_garsdc} & Surrogate & 0.05 & 3000+ & 6.0 & - \\
    GALD \cite{li_gald} & Surrogate & 0.063 & 10 & 20.6 & - \\
    RAD* \cite{chen_rad} & Surrogate & 0.063 & 10 & 27.2 & 47.2 \\
    GHFD* \cite{wang_ghfd} & Surrogate & 0.063 & 50 & 12.7 & 42.3 \\
    UEA* \cite{wei2019uea} & Surrogate & 0.063 & 50 & 28.5 & 50.7 \\
    DAG* \cite{xie2017dag} & Surrogate & 0.063 & 50 & 28.6 & 50.7 \\
    RAP* \cite{li2018rap} & Surrogate & 0.063 & 50 & 24.7 & 49.5 \\
    EBAD$^\dagger$ \cite{cai_ebad} & Victim & 0.039 & 10 & 34.9 & - \\
    AttentionFool \cite{lovisotto_attnfool} & Victim & - & 10-150 & 21.0 & - \\
    OATB \cite{leng_oatb} & Victim & 0.078 & 20 & 26.6 & - \\
    DBA \cite{lian_dba} & Victim & - & 50 & -  & 56.7 \\
    \rowcolor{lightgray} \textbf{AFOG} & \textbf{Victim} & \textbf{0.031} & \textbf{10} & \textbf{4.1} & \textbf{7.3} \\
    \bottomrule
  \end{tabular}
  \label{tab:related_comparisons}
\end{table}
\egroup

\bgroup
\addtolength{\tabcolsep}{-0.05em}
\begin{table*}[h]
  \fontsize{7pt}{7pt}\selectfont
  \centering
  \caption{Timing and Imperceptibility Results. $L_2$ represents the average $L_2$ norm difference between perturbed and clean images, $L_0$ is the average proportion of perturbed pixels, SSIM is the structural similarity index measure, $\mu_\Delta$ is the average perturbation magnitude, and t is average total attack time for all ten iterations in seconds.}
  \begin{tabular}{cccccc|ccccc|ccccc}
    \toprule
     & \multicolumn{5}{c}{AFOG} & \multicolumn{5}{c}{AFOG-V} & \multicolumn{5}{c}{AFOG-F}  \\
    \cline{3-5}\cline{8-10}\cline{13-15}
    Model & \rule[1pt]{0pt}{7pt}$L_2$ & $L_0$ & SSIM & $\mu_\Delta$ & time & $L_2$ & $L_0$ & SSIM & $\mu_\Delta$ & time & $L_2$ & $L_0$ & SSIM & $\mu_\Delta$ & time \\
    \midrule
    DETR-R50 \cite{carion2020} & 0.0322 & 0.9707 & 0.8715 & 0.0173 & 1.45 & 0.0323 & 0.9707 & 0.8716 & 0.0172 & 0.99 & 0.0323 & 0.9710 & 0.8717 & 0.0172 & 1.21 \\
    DETR-R101 \cite{carion2020} & 0.0323 & 0.9707 & 0.8721 & 0.0173 & 1.47 & 0.0323 & 0.9706 & 0.8724 & 0.013 & 1.16 & 0.0323 & 0.9708 & 0.8724 & 0.0172 & 1.70\\
    Deform.-DETR \cite{zhu2020deformable} & 0.0323 & 0.9719 & 0.8711 & 0.0174 & 1.63 & 0.0323 & 0.9713 & 0.8716 & 0.0173 & 1.63 & 0.0012 & 0.9714 & 0.8717 & 0.0173 & 1.87 \\ 
    R50 \cite{he2016resnet} & 0.0317 & 0.9658 & 0.8343 & 0.0171 & 2.70 & 0.0317 & 0.9650 & 0.8348 & 0.0170 & 2.34 & 0.0317 & 0.9654 & 0.8346 & 0.0170 & 3.16\\
    AlignDETR \cite{cai2023aligndetr} & 0.0319 & 0.9657 & 0.8347 & 0.0170 & 2.41 & 0.0319 & 0.9646 & 0.8349 & 0.0170 & 2.33 & 0.0319 & 0.9647 & 0.8349 & 0.0170 & 3.29 \\ 
    ViTDet \cite{li2022vitdet} & 0.0318 & 0.9671 & 0.8353 & 0.0171 & 6.88 & 0.0318 & 0.9657 & 0.8361 & 0.0170 & 6.67 & 0.0318 & 0.9664 & 0.8355 & 0.0171 & 7.33  \\
    ConvNext \cite{liu2022convnext} & 0.0318 & 0.9666 & 0.8342 & 0.0171 & 5.38 & 0.0318 & 0.9654 & 0.8349 & 0.0170 & 5.26 & 0.0318 & 0.9663 & 0.8347 & 0.0170 & 5.86 \\
    Swin-L \cite{liu2021Swin} & 0.0327 & 0.9724 & 0.8673 & 0.0175 & 7.13 & 0.0327 & 0.9716 & 0.8680 & 0.0173 & 7.28 & 0.0327 & 0.9722 & 0.8678 & 0.0174 & 8.99 \\
    InternImage \cite{wang2022internimage} & 0.0318 & 0.9665 & 0.8360 & 0.0170 & 6.35 & 0.0318 & 0.9653 & 0.8367 & 0.0170 & 6.23 & 0.0318 & 0.9660 & 0.8364 & 0.0171 & 6.70 \\ 
    FocalNet \cite{yang2022focal} & 0.0320 & 0.9665 & 0.8365 & 0.0172 & 8.96 & 0.0320 & 0.9657 & 0.8378 & 0.0171 & 8.84 & 0.0320 & 0.9659 & 0.8374 & 0.0171 & 9.74\\
    EVA \cite{EVA} & 0.0370 & 0.9666 & 0.8240 & 0.0172 & 54.34 & 0.0370 & 0.9665 & 0.8246 & 0.0171 & 54.53 & 0.0370 & 0.9665 & 0.8242 & 0.0171 & 51.18 \\
    DETA \cite{ouyangzhang2022nms} & 0.0481 & 0.9662 & 0.8096 & 0.0170 & 13.20 & 0.0481 & 0.9654 & 0.8099 & 0.0170 & 13.10 & 0.0481 & 0.9654 & 0.8099 & 0.0170 & 13.13\\
    \bottomrule
  \end{tabular}
  \label{tab:metrics}
\end{table*}
\egroup

Figure~\ref{fig:examples} shows six visual examples and the comparison of DETR results for benign detection (Row 1) and adversarial detection under the AFOG attack (Row 2) scenarios. The first example shows that DETR detects the person and surfboard correctly under no attack (Row 1) but outputs erroneous detections under the AFOG attack (Row 2). Additional visual examples for all twelve transformers are available in the Supplementary Material.
 
We report the distortion effect and timing cost for each of the 12 transformer models in Table~\ref{tab:metrics}. We note several interesting observations: (i) Considering the $L_2$ and SSIM metrics, AFOG attacks induce very similar levels of distortion for 10 out of 12 transformer models. The two largest detection transformers, DETA~\cite{ouyangzhang2022nms} and EVA~\cite{EVA}, show worse $L_2$ and SSIM scores for AFOG, AFOG-V and AFOG-F. (ii) Considering the $L_0$ metric, only DETA shows the lowest $L_0$ value consistently for AFOG, AFOG-V and AFOG-F. (iii) As expected, the time required to attack a detection transformer model is strongly correlated to the number of model parameters. For instance, EVA~\cite{EVA} has over 1 billion parameters, and took much longer on average to attack an input image over 10 iterations compared to the other 11 models. (iii) DETA~\cite{ouyangzhang2022nms} has 218.8 million parameters and is similar to other models within the Detrex framework: FocalNet~\cite{yang2022focal}, InternImage~\cite{wang2022internimage}, Swin-L~\cite{liu2021Swin} and ConvNext~\cite{liu2022convnext}. However, we observe that it consistently takes about $1.5-2\times$ longer time to attack one input image on average with our AFOG attack or its special case extensions AFOG-V or AFOG-F. (iv) AFOG, AFOG-V and AFOG-F have similar average perturbation magnitude.    

\subsection{Effectiveness on CNN-Based Detectors}
\noindent
This section compares AFOG with four attacks designed for CNN-based models: TOG~\cite{chow2020tog}, UEA~\cite{wei2019uea}, RAP~\cite{li2018rap}, and DAG~\cite{xie2017dag}. They are surrogate-based attacks, and a majority of them (e.g., DAG, RAP, UEA) can only directly attack two-stage detectors (exemplified by Faster-R-CNN~~\cite{ren2016frcnn}) and rely on adversarial transferability to attack single-stage CNN-based detectors like YOLOv3~\cite{redmon2018yolov3} and SSD~\cite{wei2016ssd}. 
Table~\ref{tab:comparison} reports the comparison results. We make two observations: (i) AFOG and TOG are the only two victim-based attacks against single-stage detectors, represented by YOLOv3 and SSD. Both AFOG and TOG can drastically reduce the benign mAP of YOLOv3 ($83.43\%$) and SSD ($76.11\%$). For YOLOv3, TOG is a stronger attack, achieving mAP of $0.56$ compared to mAP of AFOG being $2.62$. However, for SSD-300, AFOG is a stronger attack, achieving mAP of $0.50$, compared to mAP of TOG ($0.86$). (ii) For two-stage CNN-based object detectors, exemplified by Faster-R-CNN, AFOG outperforms all four other victim-based attacks, achieving the lowest adversarial mAP score of $2.07\%$. (iii) AFOG also uses the same distortion magnitude budget $L_\infty$ as TOG, while achieving higher attack success rate (lower mAP) on SSD-300 and FRCNN with decreased $L_2$ distortion cost. Hence, AFOG excels at attacking both transformer-based detectors and CNN-based detectors. 

\bgroup
\addtolength{\tabcolsep}{-0.1em}
\begin{table}[h]
    \fontsize{7pt}{7pt}\selectfont
    \setlength{\columnwidth}{1pt}
    \centering
     \caption{Comparing AFOG with four state-of-the-art attacks on representative CNN-based object detectors. mAPs of existing attacks were taken from respective papers \cite{chow2020tog}. (-) indicates N/A.}
    \begin{tabular}{cccccccc}
    \toprule
    & & & & \multicolumn{4}{c}{Distortion Cost} \\
    \cline{5-8}
    Model & Attack & mAP & t &\rule[1pt]{0pt}{7pt}$L_\infty$ & $L_2$ & $L_0$ & SSIM \\
    \midrule
    \multirow{3}{*}{YOLOv3} & Benign & 83.43 & 0.0 & 0.0 & 0.0 & 0.0 & 1.0 \\
    & TOG \cite{chow2020tog} & \textbf{0.56} & \textbf{0.98} & 0.031 & 0.083 & 0.984 & \textbf{0.875} \\
    & AFOG & 2.28 & 1.31 & 0.031 & \textbf{0.013} & \textbf{0.855} & 0.801 \\
    \midrule
   \multirow{5}{*}{SSD-300} & Benign  & 76.11 & 0.0 & 0.0 & 0.0 & 0.0 & 1.0 \\
    & UEA \cite{wei2019uea}  & 20.0 & - & - & - & - & - \\
    & DAG \cite{xie2017dag}  & 64.0 & - & - & - & - & - \\ 
    & TOG \cite{chow2020tog} & 0.86 & \textbf{0.39} & 0.031 & 0.120 & 0.975 & \textbf{0.879} \\
    & AFOG & \textbf{0.50} & 0.49 & 0.031 & \textbf{0.022} & \textbf{0.858} & 0.793 \\
    \midrule
    \multirow{6}{*}{FRCNN} & Benign & 67.37 & 0.0 & 0.0 & 0.0 & 0.0 & 1.0 \\
    & UEA \cite{wei2019uea} & 5.0 & 0.17 & 0.343 & 0.191 & 0.959 & 0.652 \\
    & RAP \cite{li2018rap} & 4.78 & 4.04 & 0.082 & 0.010 & 0.531 & 0.994 \\ 
    & DAG \cite{xie2017dag} & 3.56 & 7.99 & \textbf{0.024} & \textbf{0.002} & \textbf{0.493} & \textbf{0.999} \\ 
    & TOG \cite{chow2020tog} & 2.64 & \textbf{1.68} &  0.031 & 0.058 & 0.976 & 0.862 \\
    & AFOG & \textbf{2.38} & 2.11 & 0.031 & 0.019 & 0.854 & 0.788 \\
    
    \bottomrule
    \end{tabular}
    \label{tab:comparison}
\end{table}
\egroup

\noindent

\subsection{Effect of Learnable Self-Attention}
\begin{figure}[h]
    \centering
    \includegraphics[width=\columnwidth]{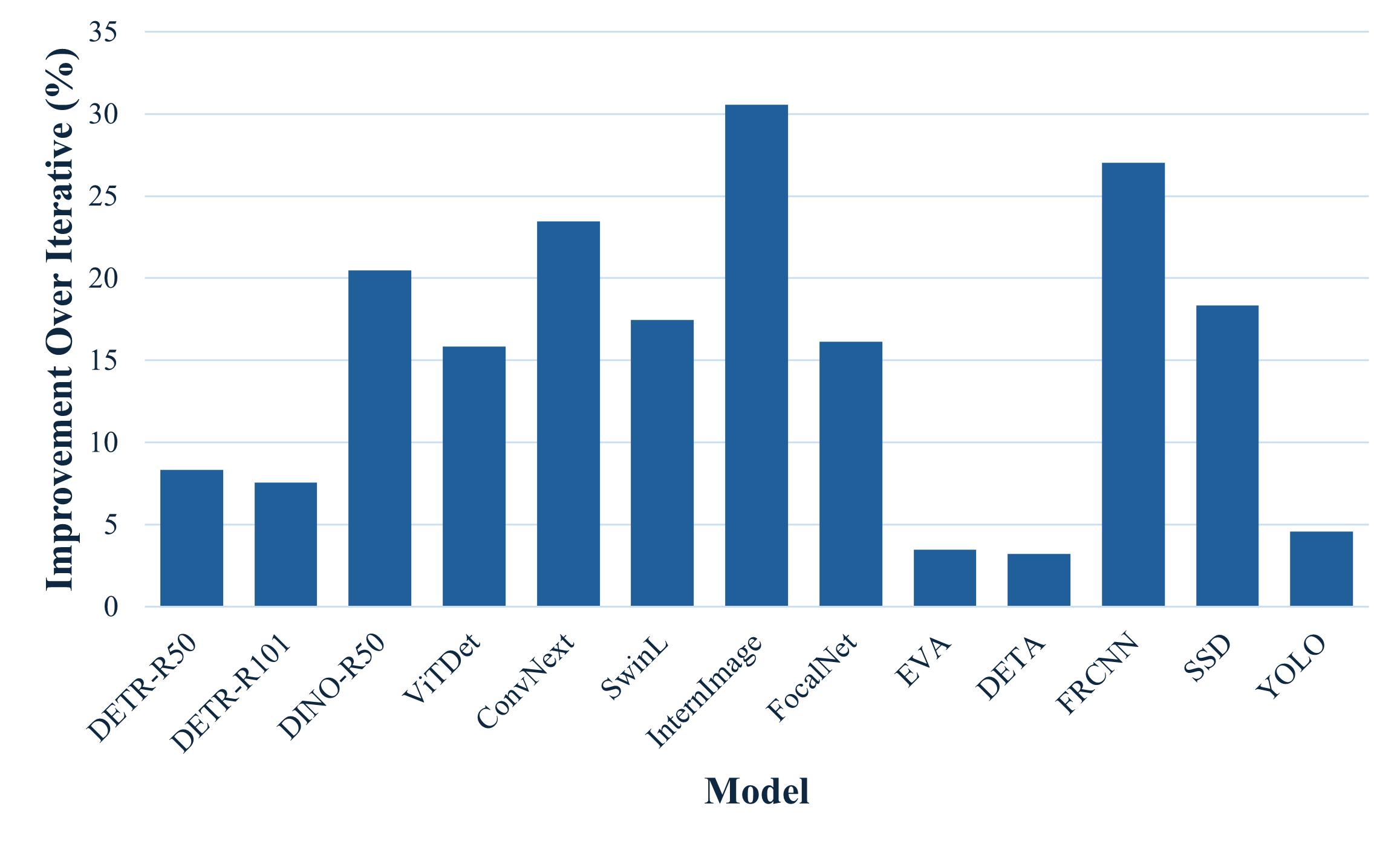}
    \caption{Improvement (\% difference) of AFOG with attention over AFOG without attention. The learnable attention mechanism improves performance by as much as 30.6\% on InternImage, with an average improvement of 15.1\% across all models.}
    \vspace{-6pt}
    \label{fig:ablation}
    \vspace{-8pt}
\end{figure}

\noindent 
We isolate and analyze the role of the learnable self-attention mechanism in empowering AFOG to successfully corrupt transformer-based models by probing for weak spots and performing fast and human-imperceptible perturbations.  Figure \ref{fig:ablation} shows a comparison between AFOG with and without its learnable attention mechanism, measured by percent difference in adversarial mAP scores. Ablation results show that our learnable attention mechanism improves performance by up to 30.6\% on InternImage, with an average of 15.1\% improvement across all models. 

\begin{figure}[h]
    \centering
    \includegraphics[width=\columnwidth]{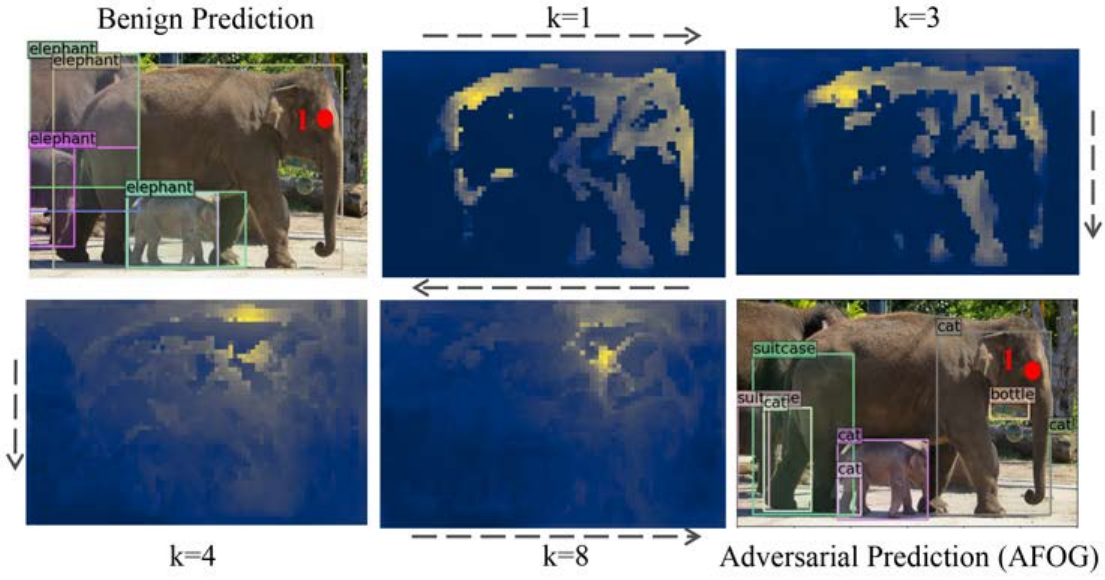}
    \caption{DETR self-attention weights analysis. Each blue frame shows the self-attention weights of DETR-R50's last encoder layer at the red indicated point in the image at a given attack iteration $k$. In early iterations, the model understands a strong association between the indicated red point on the elephant and the rest of the elephant shape. As the attack iterates (grey arrows), this association becomes corrupted, evidenced by the lack of encoder self-attention structure in later stages. By the $8^{th}$ iteration, the selected point has lost association with the rest of the elephant object, demonstrating that AFOG has disrupted learned knowledge.}
    \vspace{-6pt}
    \label{fig:selfattention}
    \vspace{-8pt}
\end{figure}

Figure~\ref{fig:selfattention} shows the self-attention map for the last encoder layer of DETR~\cite{carion2020} at initialization and three progressive attack iterations ($3^{rd}, 4^{th}$, and $8^{th}$). We observe that, as the attack progresses in iterations (indicated by grey arrows), the self-attention map becomes increasingly less associated with the benign detected objects. By the $8^{th}$ iteration of the AFOG attack, the self-attention maps have become almost entirely disjoint from the benign detections of the victim model (DETR) prior to the attack. We note that, when the model's self-attention map that reflects learned detection knowledge is changed, it is unlikely for prior knowledge to be kept intact. During the AFOG attack, the perturbed inputs focused by AFOG's own adversarial attention corrupt the pixel-level relationships learned by a well-trained victim model prior to adversarial injection (e.g., $k=1$), causing catastrophic forgetting to occur. 

\subsection{Worst Case Analysis}
\begin{figure}[h]
    \centering
    \vspace{-3pt}
    \includegraphics[width=\columnwidth]{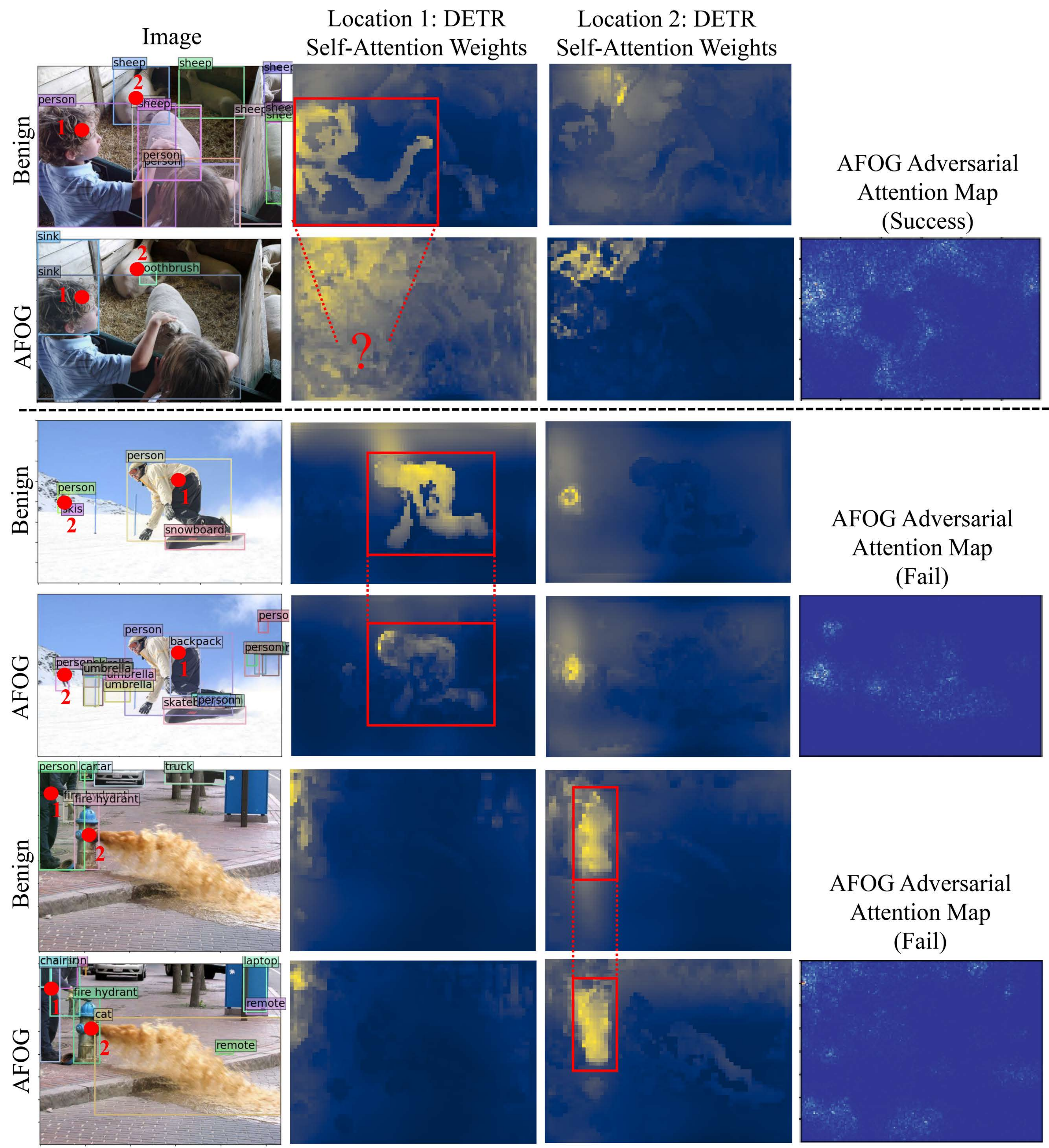}
    \caption{Worst case analysis with DETR-R50 as victim model. Rows 1-2 show a benign image and the successful AFOG perturbed image. Rows 3-6 show two cases where AFOG attack did not succeed. Columns 2-3 show DETR's encoder's self-attention map for the two locations marked in red in column one. Column 4 shows AFOG's internal attention map, where white pixels indicate high attention and blue indicates low. In the successful case (Rows 1-2), encoder self-attention maps (Cols. 2-3) lose associative structure. In the unsuccessful cases (Rows 4 and 6), AFOG fails to disrupt this structure and predictions are preserved. AFOG adversarial attention maps (Col. 4) show focus on important object structure in the successful case (Row 2) and lack of focus in the unsuccessful cases (Rows. 4, 6). }
    \vspace{-8pt}
    \label{fig:fails}
\end{figure}

\vspace{-3pt}
\noindent
Our experimental results on 12 transformer-based detectors and 3 representative CNN-based detectors have shown that AFOG is a fast and effective attack. However, there are still cases where AFOG fails in its adversarial perturbations within a total of 10 iterations, one setting of our attack termination hyper-parameter. By investigating those cases where AFOG did not succeed, we found that the most common situation is where the attack fails to corrupt 
the central subject of the victim input image, e.g., a person. Figure \ref{fig:fails} gives three visual examples, with the top one (Rows 1-2) being a successful attack followed by two failed cases (Rows 3-6) when attacking DETR-R50. Columns 2-3 in all six rows show DETR's encoder's last layer's self-attention map for the first and second locations indicated by red points in the first column. Rows 1, 3, and 5 show benign situations and their corresponding self-attention maps, and Rows 2, 4, 6 show the self-attention maps after the final attack iteration. The fourth column shows AFOG's adversarial attention map after the final attack iteration.

We make three observations: (i) For the successful attack case (Rows 1-2), the self-attention maps of victim model DETR-R50 lose association between the indicated red point and the rest of the foreground object it comprises. (ii) In both unsuccessful cases (Rows 3-6), AFOG fails to disrupt this association. (iii) Consider AFOG's adversarial attention maps from the successful case in Figure \ref{fig:fails} (Row 2, Col. 4), we observe a clear focus on the key objects in the victim input image. In comparison, AFOG's adversarial attention maps for the two failure cases (Row 4, Col. 4 and Row 6, Col. 4) both fail to focus on foreground objects. The first failure case (Rows 3 and 4) shows that the AFOG attention map attributes significant weight to the distant person in the left of the victim image, resulting in a large number of fabricated predictions in that region. Similarly, the attention map for the second victim image (Rows 5 and 6) appears to entirely miss the fire hydrant, resulting in a failure to disrupt this object. In these cases, AFOG's adversarial attention has learned to focus on the wrong pixels.

\section{Conclusion}
\noindent 
We have presented AFOG, an Attention-Focused Offensive Gradient attack. AFOG is effective for attacking both advanced object detection transformers and traditional CNN-based detectors with a unified, architecture-agnostic framework. AFOG utilizes a learnable attention mechanism to enable its adversarial perturbations to focus on vulnerable areas of images in multi-box detection tasks. AFOG's attack loss function integrates multiple feature losses (e.g., bounding-box loss, class loss) through learnable attention updates with iterative injection of adversarial perturbations. Finally, AFOG is efficient and stealthy. Adversarial perturbations generated by AFOG are visually imperceptible and yet can cause well-trained detectors to fail miserably. Extensive experiments on state-of-the-art object detectors show that AFOG is consistently effective across twelve object detection transformers. Comparative evaluation with nearly a dozen SOTA methods shows that AFOG significantly outperforms surrogate-based and victim-based attacks on both object detection transformers and CNN-based object detectors.
\\
\\
{\bf Acknowledgement.\/}
This research is partially sponsored by the NSF CISE grants 2302720 and 2312758, CISCO Edge AI program, and GTRI PhD Fellowship. 

{\small
\bibliographystyle{ieee_fullname}
\bibliography{iccv2023AuthorKit/sources}
}

\maketitlesupplementary
\setcounter{section}{0}
\section{Reproducibility Statement}
\label{sec:rep_statement}
\noindent
We make the following efforts to enhance the reproducibility of our results.

\begin{itemize}

    \item For AFOG's implementation a link to an anonymous downloadable source repository is included in our abstract. The source includes links for all datasets and models used in our experiments.

    \item Our experiment details are given in Section 4, containing selected hyperparameters and hardware specifications.

    \item We also show example images under benign and AFOG attack scenarios throughout the paper and appendix.

\end{itemize}

\section{Additional Experimental Setup}

\label{sec:models}

\subsection{Model Details}
\noindent
In our experiments to validate the effectiveness of AFOG attacking vision transformer models for object detection, we select twelve transformers of varying model sizes, ranging from Detection Transformer (DETR) \cite{carion2020}, a lightweight 40 million parameter model, to EVA \cite{EVA} with more than one billion parameters. We use the Detrex framework \cite{ren2023detrex}, built on top of Detectron2 \cite{wu2019detectron2}, and DINO \cite{zhang2022dino} to standardize our model implementations. Both DINO and Detrex adapt numerous general-purpose transformer models to object detection. Further, to demonstrate that AFOG also works beyond Detrex, we choose some transformer models such as DETR and Swin from their original repositories associated with the original papers instead of the versions provided in Detrex. In this section of the Supplementary Material, we first provide a brief overview of each of the twelve transformer models.

\vspace{3pt}
\noindent
\textbf{Detection Transformer (DETR)}\cite{carion2020} adapts the transformer architecture for object detection framed as a set prediction task. It implements a transformer encoder-decoder architecture and a global set loss with bipartite matching. We explore DETR with ResNet-50 and ResNet-101 backbones \cite{he2016resnet}. DETR is the foundation of many other detection transformers \cite{cai2023aligndetr, zhu2020deformable, zhang2022dino}, making it a compelling target for AFOG. Original code adapted for our implementation is available at \url{https://github.com/facebookresearch/detr}

\vspace{3pt}
\noindent
\textbf{Deformable-DETR}\cite{zhu2020deformable} improves upon DETR's attention mechanism by confining self-attention to a small region around a given point. This matches DETR's performance in significantly reduced training time and improves performance on small object detection.  We use the version trained by Detrex with a ResNet-50 backbone. It is pretrained on ImageNet-1k and then trained on COCO 2017. Attacking Deformable-DETR demonstrates that AFOG is effective against its unique self-attention mechanism. Original code adapted for our implementation is available at \url{https://github.com/fundamentalvision/Deformable-DETR}

\vspace{3pt}
\noindent
\textbf{DINO} \cite{zhang2022dino} advances DETR with improved denoising anchor boxes. DINO supports numerous backbones, making it ideal for attacking many types of models with AFOG. We choose to attack DINO to assess the viability of AFOG against a variety of backbones and to investigate AFOG's performance under robust denoising. Original DINO code adapted for our implementation is available at \url{https://github.com/IDEA-Research/DINO}

\vspace{3pt}
\noindent
\textbf{AlignDETR} \cite{cai2023aligndetr} addresses an alignment issue in DETR that incorrectly matches correct predictions with the wrong ground-truths during training. The authors remedy this problem with an IoU-aware binary cross entropy (BCE) loss, mixed-matching, and a sample weighting method that reduces the effects of unimportant samples. We attack AlignDETR to explore AFOG's performance against this alternative loss functon approach. Our implementation uses a ResNet-50 backbone, and it is trained on COCO 2017.
Original code adapted for our implementation is available at \url{https://github.com/IDEA-Research/detrex/tree/main/projects/align_detr}

\vspace{3pt}
\noindent
\textbf{ViTDet} \cite{li2022vitdet} adapts the original vision transformer (ViT) to object detection. We use the version trained by Detrex, which uses a plain ViT as the backbone for ViTDet with masked auto encoder and ImageNet-1k pretraining. ViTDet is further used as a backbone for DINO, and the combination architecture is trained on COCO 2017. Like DINO and DETR, ViT is a widely popular model that has been adapted for numerous uses. We choose ViTDet for our experiments to investigate AFOG's potential applicability to all ViT-based models. Original code adapted for our implementation is available at \url{https://github.com/IDEA-Research/detrex/tree/main/projects/dino}

\vspace{3pt}
\noindent
\textbf{ConvNeXt-Large-384} \cite{liu2022convnext} ConvNeXt is a modernized pure convolutional neural network that is designed to compete with transformers on object detection and semantic segmentation tasks. We use ConvNeXt-Large-384 pretrained by Detrex on Imagenet-22k. ConvNeXt is used as a backbone with the DINO transformer, and both are trained on COCO 2017. Attacking ConvNeXt demonstrates that AFOG is effective against modern CNNs as well as transformers. Original code adapted for our implementation is available at \url{https://github.com/IDEA-Research/detrex/tree/main/projects/dino}

\vspace{3pt}
\noindent
\textbf{Swin-L} \cite{liu2021Swin} is a hierarchical transformer designed for computer vision. Swin uses shifting windows to boost efficiency and limit the computational cost of its self-attention mechanism. This efficiency also makes it viable at a range of model sizes. We use Swin-Large-384-4-Scale from Detrex, pretrained on ImageNet-22k and trained on COCO 2017. We choose to attack Swin in our experiments to investigate AFOG's efficacy against its shifting window mechanism and because it is a popular backbone for a variety of object detection models. Original code adapted for our implementation is available at \url{https://github.com/IDEA-Research/DINO} 

\vspace{3pt}
\noindent
\textbf{InternImage-Large} \cite{wang2022internimage} is a CNN-based foundation model that leverages deformable convolution instead of sparse kernels. This reduces the inductive bias of traditional CNNs and enables InternImage to learn larger-scale patterns from larger datasets. We use InternImage as a backbone for the DINO transformer. The pairings are pretrained on Imagenet-22k and trained on COCO 2017. We choose InternImage-Large for our experiments to explore AFOG's applicability to deformable convolution. A larger version of InternImage is also currently one of the strongest models on the COCO object detection leaderboard \cite{cocoleaderboard}. We use InternImage-Large, a deeper model with more parameters than vanilla InternImage, because of its stronger benign performance. Original code adapted for our implementation is available at \url{https://github.com/IDEA-Research/detrex/tree/main/projects/dino}

\vspace{3pt}
\noindent
\textbf{FocalNet} \cite{yang2022focal} replaces self-attention with a focal modulation mechanism that describes interactions of vision tokens. This mechanism is comprised of three parts: hierarchical contextualization, gated aggregation, and element-wise modulation. Using FocalNet in our experiments demonstrates AFOG's viability against this focal modulation mechanism, which may have different weaknesses than the traditional self-attention inside a transformer. We adapt FocalNet-Large-4scale, pretrained on ImageNet-22k for object detection with the DINO transformer. This combination is trained on COCO 2017. Original code adapted for our implementation is available at \url{https://github.com/IDEA-Research/detrex/tree/main/projects/dino}

\vspace{3pt}
\noindent
\textbf{EVA} \cite{EVA} is a large vision foundation model. It is based on a classical ViT that is pre-trained on masked vision-text features. It demonstrates strong performance in various transfer learning tasks. We adapt EVA for object detection with DINO and Detrex. The Detrex version of EVA is an EVA-01 model trained on COCO 2017 with additional large-scale jittering (LSJ) augmentation. We select EVA for our experiments to investigate AFOG's performance against a large foundation model with extensive pre-training. 
Original code adapted for our implementation is available at \url{https://github.com/IDEA-Research/detrex/tree/main/projects/dino_eva}

\vspace{3pt}
\noindent
\textbf{DETA} \cite{ouyangzhang2022nms} reintroduces IoU-based assignment and non-maximum suppression (NMS) to the classic DETR architecture. The authors show that this is superior to the bipartite matching used in the original DETR. Our version of DETA is implemented via Detrex and uses a Swin-Large-384 backbone. It is pretrained on ImageNet-1k and trained on COCO 2017. We include DETA with a Swin backbone in our experiments to investigate AFOG's performance against IoU-based assignment and NMS, and also to explore the Swin backbone with another architecture in addition to DINO. Original code adapted for our implementation is available at \url{https://github.com/IDEA-Research/detrex/tree/main/projects/dino_eva}

\vspace{6pt}
\noindent
In addition to detection transformers, we also evaluate AFOG attacking conventional CNN-based object detection models, represented by Faster-R-CNN, SSD and YOLOv3.

\vspace{3pt}
\noindent
\textbf{Faster R-CNN} (FRCNN) \cite{ren2016frcnn} is a two-phase CNN-based object detector that uses a region proposal network to share convolutional features with the detection network. It is the standard victim object detector in many other attacks \cite{wei2019uea, chow2020tog, xie2017dag, li2018rap} making it an ideal standard of comparison.  We use a version with a ResNet-50 backbone, trained on Pascal VOC 2007. Original code adapted for our implementation is available at \url{https://github.com/chenyuntc/simple-faster-rcnn-pytorch}

\vspace{3pt}
\noindent
\textbf{YOLOv3} \cite{redmon2018yolov3} is a single-phase CNN-based object detector that predicts existence of objects, their bounding boxes and their class labels in a single pass. YOLOv3 is a version of YOLO family of algorithms with improved performance, and we used the pre-trained YOLOv3 on Pascal VOC 2007. YOLOv3 is another popular model among other attacks, making it ideal for comparison. Original code adapted for our implementation is available at \url{https://github.com/qqwweee/keras-yolo3}

\vspace{3pt}
\noindent
\textbf{SSD-300} \cite{wei2016ssd} is also a single-phase CNN-based object detector, which divides an image into default regions at varying scales and aspect ratios and predicts object existence, bounding box and class label with confidence scores for each bounding box. High-scoring boxes are also adjusted to better fit predicted objects. We use a pretrained SSD-300 on Pascal VOC 2007, which takes 300$\times$300 inputs. Original code adapted for our implementation is available at \url{https://github.com/pierluigiferrari/ssd_keras}

\subsection{Hyperparameter Tuning}
\begin{figure}[h]
    \includegraphics[width=\columnwidth]{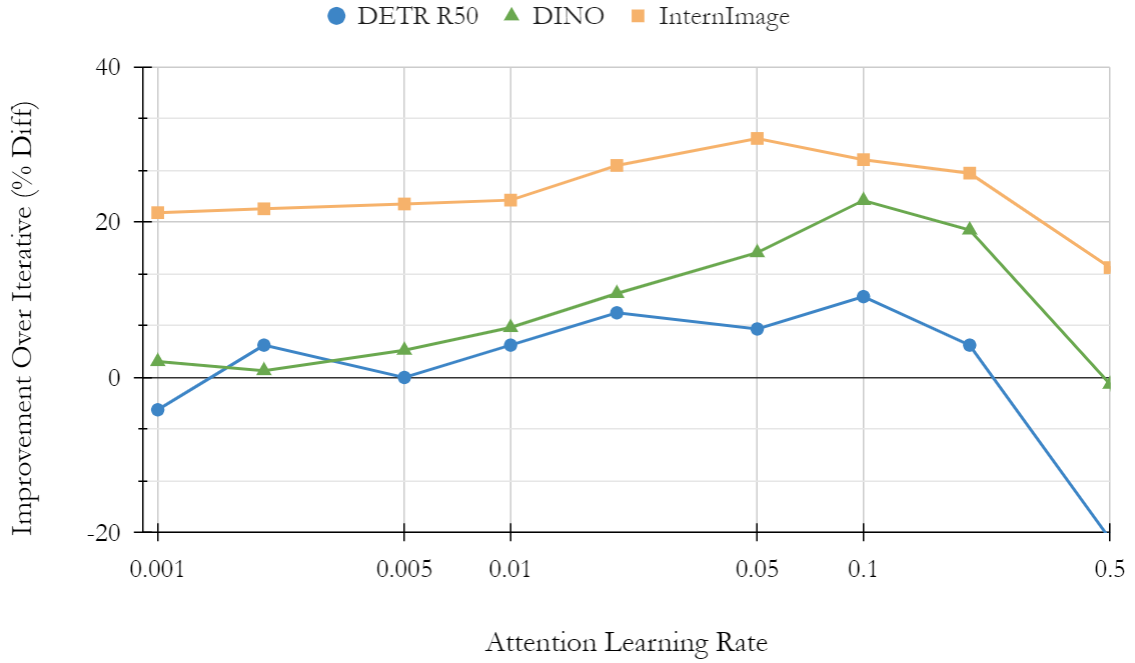}
    \caption{AFOG improvement over iterative for varying values of attention learning rate across three detection transformers.}
    \label{fig:attn_lr_tuning}
\end{figure}

\noindent
We introduce an attention learning rate hyperparameter (see Equation \ref{eq:backprop_a}) that controls the sensitivity of AFOG's adversarial attention maps to gradient updates. Figure \ref{fig:attn_lr_tuning} shows the performance of AFOG with varying attention learning rates compared to AFOG without attention (attention learning rate zero) for three detection transformers. We observe that an attention learning rate of 0.1 shows the strongest performance across all three models, and so we use this value for all models and all experiments.

%

\section{AFOG Variants and Pseudocode}

In addition to the pseudocode for AFOG provided in our main paper, in this section, we also provide the pseudocode for AFOG-V in Algorithm \ref{alg:afogv} and AFOG-F in Algorithm \ref{alg:afogf}. 

The key difference between AFOG and AFOG-V is the replacement of $O_x$ with a set of zero predictions $\varnothing$ instead of forward propagating image $\mathcal{O}_x \leftarrow f_D(x; \vartheta)$. AFOG-V also aims to minimize $\mathcal{L}_{AFOG}$ according to its slightly different formulation. Algorithm~\ref{alg:afogv} provides a sketch of the pseudo code.

AFOG-F changes AFOG by modifying benign predictions $\mathcal{O}_f$ to change all confidence scores to 1.0. This means every benign prediction by the model, regardless of quality, is used as a ground truth during attack iteration. AFOG-F aims to minimize $\mathcal{L}_{AFOG}$, learning perturbations that increase the likelihood of inducing these spurious low-quality predictions. This is reflected in Algorithm \ref{alg:afogf}.

\begin{algorithm}[ht!]
\caption{AFOG-V attack}\label{alg:afogv}
\begin{algorithmic}[1]
\Require 
 Victim image $x\in \mathcal{D}$, test-set $\mathcal{D}$, 
Victim pre-trained model $f_D(\vartheta)$,
Perturbation step size $\alpha_P$, 
Attention step size $\alpha_A$, 
Num. iters $T$, 
Max pert. $\epsilon$.
\State Initialize $\mathcal{O}_x \leftarrow \varnothing$
\State Initialize attention map $A_0\leftarrow 1$;
\State Initialize perturbation $P_0\leftarrow Random(-\epsilon, \epsilon)$;
\State Initialize step variable $k \gets 1$;
\While{$k \leq T$}
    \State Attack image ${x_{adv}}_{k} \gets \prod\limits_{x, \epsilon}(x + A_{k} \odot P_{k}$);
    \State Forward propagate $x_{adv}$ through $f_{D}(\vartheta, x_{adv})$;
    \State Compute bbox-loss $ \mathcal{L}_{bbox}(x_{adv}$, $\mathcal{O}_x; \vartheta)$;
    \State Compute cls-loss $\mathcal{L}_{cls}(x_{adv}$, $\mathcal{O}_x; \vartheta)$; 
\State $\mathcal{L}_{AFOG}(x_{adv}, \mathcal{O}_x; \vartheta)=$ $-$bbox-loss $-$ cls-loss;
    \State Calculate losses with respect to $A_k$ and $P_k$: $\mathcal{L}_A(x_{adv}, \mathcal{O}_x; \vartheta), \mathcal{L}_P(x_{adv}, \mathcal{O}_x; \vartheta)$;
    \State Normalize attention loss $\mathcal{L}_A \gets$ Norm$(\mathcal{L}_A)$;
    \State Take sign of perturbation loss $\mathcal{L}_P \gets$ Sign$(\mathcal{L}_P)$;
    \State $A_{k+1} \gets A_k - \alpha_A\mathcal{L}_A$; 
    \State $P_{k+1} \gets P_k - \alpha_P\mathcal{L}_P$;
    \State $k \gets k+1$;
\EndWhile
\State ${x_{adv}}_{k+1} \gets \prod\limits_{x, \epsilon}(x + A_{k+1} \odot P_{k+1})$;
\State \Return $x_{adv}$
\end{algorithmic}
\end{algorithm}
\vspace{8pt}
\begin{algorithm}[ht!]
\caption{AFOG-F attack}\label{alg:afogf}
\begin{algorithmic}[1]
\Require 
 Victim image $x\in \mathcal{D}$, test-set $\mathcal{D}$, 
Victim pre-trained model $f_D(\vartheta)$,
Perturbation step size $\alpha_P$, 
Attention step size $\alpha_A$, 
Num iters $T$, 
Max pert $\epsilon$.
\State Initialize $\mathcal{O}_f \leftarrow f_D(x; \vartheta)$
\State Initialize attention map $A_0\leftarrow 1$;
\State Initialize perturbation $P_0\leftarrow Random(-\epsilon, \epsilon)$;
\State Initialize step variable $k \gets 1$;
\While{$k \leq T$}
    \State Attack image ${x_{adv}}_{k} \gets \prod\limits_{x, \epsilon}(x + A_{k} \odot P_{k}$);
    \State Forward propagate $x_{adv}$ through $f_{D}(\vartheta, x_{adv})$;
    \State Compute bbox-loss $ \mathcal{L}_{bbox}(x_{adv}$, $\mathcal{O}_x; \vartheta)$;
    \State Compute cls-loss $\mathcal{L}_{cls}(x_{adv}$, $\mathcal{O}_x; \vartheta)$; 
\State $\mathcal{L}_{AFOG}(x_{adv}, \mathcal{O}_x; \vartheta)=$ $-$bbox-loss $-$ cls-loss;
    \State Calculate losses with respect to $A_k$ and $P_k$: $\mathcal{L}_A(x_{adv}, \mathcal{O}_x; \vartheta), \mathcal{L}_P(x_{adv}, \mathcal{O}_x; \vartheta)$;
    \State Normalize attention loss $\mathcal{L}_A \gets$ Norm$(\mathcal{L}_A)$;
    \State Take sign of perturbation loss $\mathcal{L}_P \gets$ Sign$(\mathcal{L}_P)$;
    \State $A_{k+1} \gets A_k - \alpha_A\mathcal{L}_A$; $P_{k+1} \gets P_k - \alpha_P\mathcal{L}_P$;
    \State $k \gets k+1$;
\EndWhile
\State ${x_{adv}}_{k+1} \gets \prod\limits_{x, \epsilon}(x + A_{k+1} \odot P_{k+1})$;
\State \Return $x_{adv}$
\end{algorithmic}
\end{algorithm}

\begin{figure*}
    \centering
    \includegraphics[width=1.9\columnwidth]{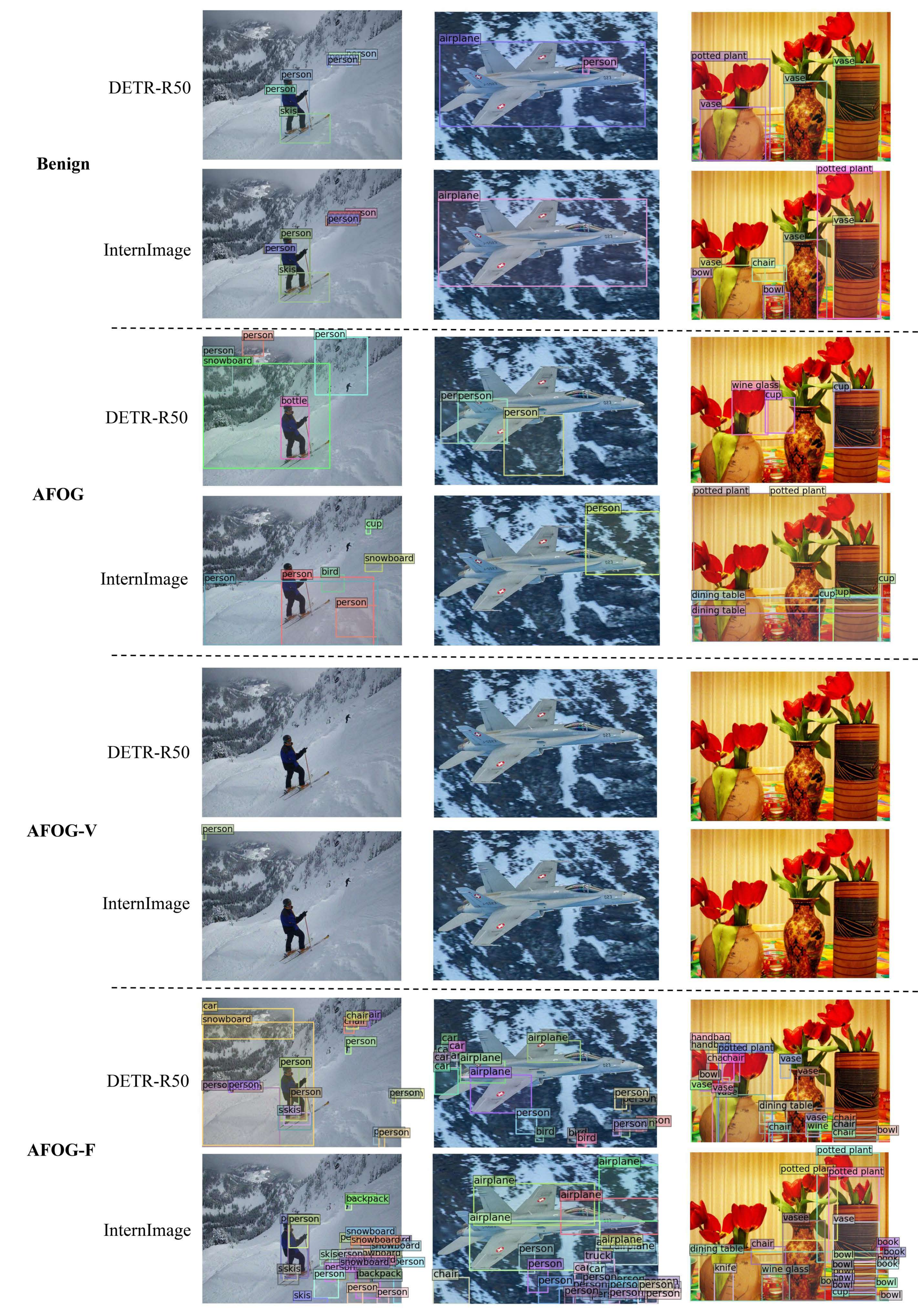}    \caption{Example images for Benign, AFOG, AFOG-V, and AFOG-F scenarios with victims R50 (DINO) and InternImage (DINO).}
    \label{fig:AFOG-vis}
\end{figure*}

\section{Adversarial Attacks: Visualization}
\noindent
In this section, we provide additional visualization of AFOG, AFOG-V, and AFOG-F attacks.

\subsection{AFOG versus AFOG-V and AFOG-F}
\noindent
Figure~\ref{fig:AFOG-vis} shows the detection results of the victim transformer detector DETR-R50 under the benign scenario for three examples from COCO testdev in Row 1. 
In Rows 2-4, we show the AFOG attack, AFOG-V attack and AFOG-F attack on these three example images. In Row 5 we show the detection results of another detection transformer model InternImage (DINO) under the benign scenario for the three example images. Rows 6-8 show the adverse effects of the AFOG attack, AFOG-V attack, and AFOG-F attack on these three example images respectively. We observe that baseline AFOG demonstrates a mixture of generating false positives, obfuscating true positives, and disrupting bounding boxes in benign cases for all three images across both models (Rows 1 and 5). AFOG-V disrupts all detections across both models, with the exception of a small "person" detection in the top left corner of Row 7, Column 1. AFOG-F induces a large quantity of spurious detections, which show different behaviors between the two models. For DETR (Row 4), the false positive detections tend to spread out across each image. For ViTDet (Row 8) the false positive detections tend to cluster in areas without foreground objects. We observe AFOG-F also fails to disrupt some true positive detections, such as the central person in Row 4, Column 1. We conclude that generic AFOG displays the positive aspects of both variants: sufficient vanishing to disrupt true positive detections paired with sufficient fabrication to induce a small number of false positives.

\subsection{Visualization of Adverse Effects of AFOG Attack on Different Detection Transformers}

\begin{figure*}[!ht]
    \centering
    \includegraphics[width=2\columnwidth]{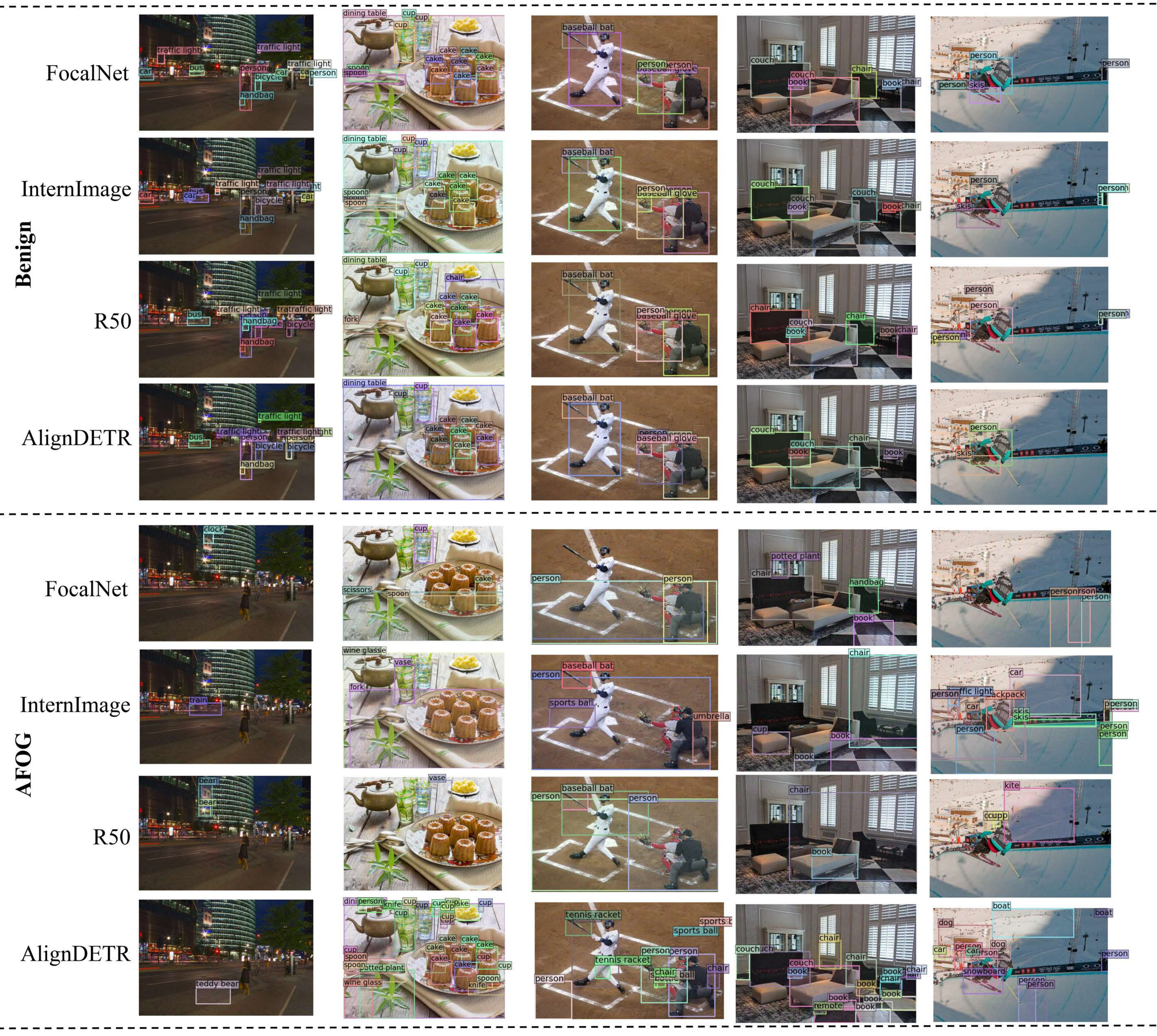}
    \caption{Comparison of Benign and AFOG-disrupted predictions on five example images with ResNet-50, InternImage, FocalNet, and Align-DETR. Despite nearly identical benign behaviors of all four models across five images, AFOG induces different malicious behavior by effectively learning unique vulnerabilities, indicating strong AFOG adaptability.}
    \label{fig:group-a-examples}
\end{figure*}

\begin{figure*}[!ht]
    \centering
    \includegraphics[width=2\columnwidth]{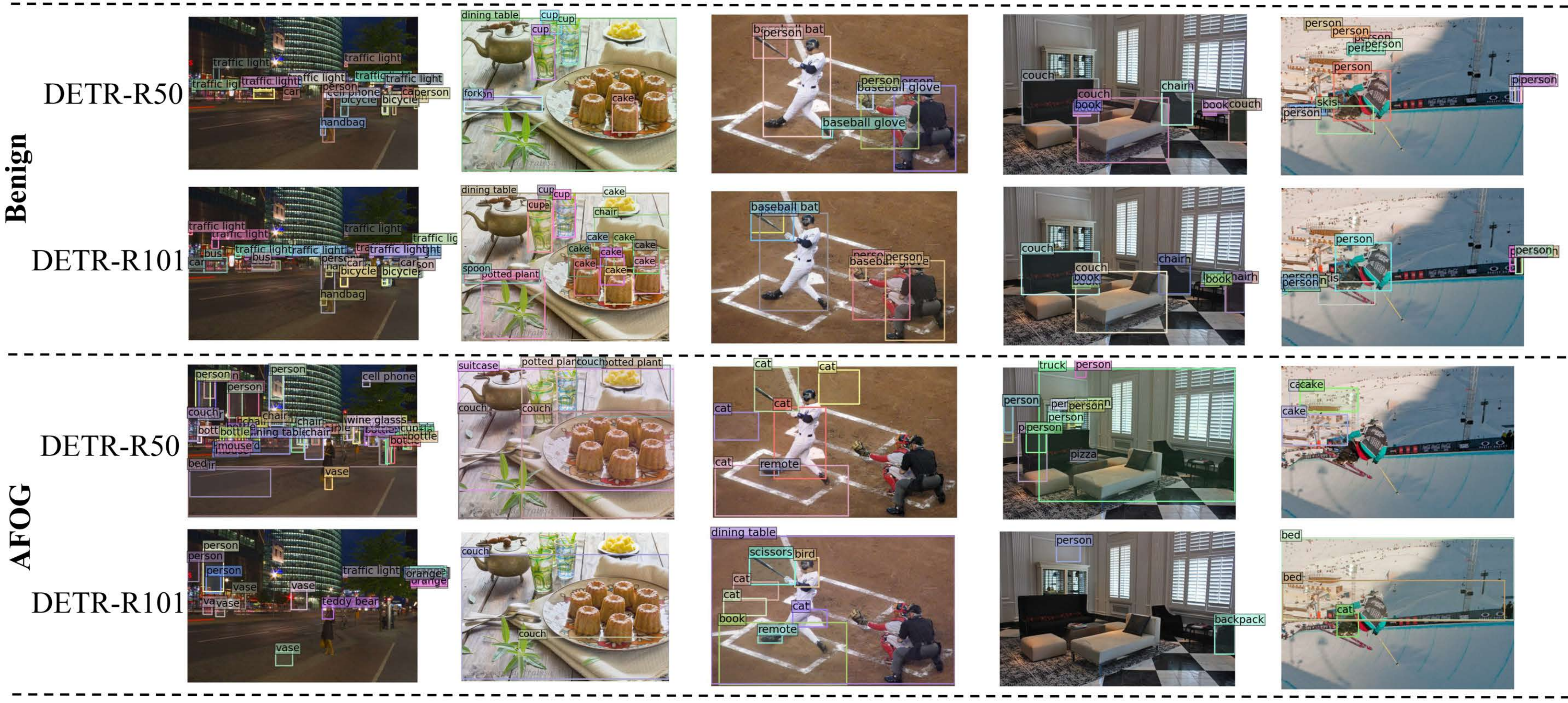}
    \caption{Comparison of Benign and AFOG-disrupted predictions on five images with two similar architectures, DETR-R50 and DETR-R101. Despite similar architectures, AFOG produces different malicious behavior by probing for unique vulnerabilities in each model.}
    \label{fig:group-b-examples}
\end{figure*}

\begin{figure*}[h]
    \centering
    \includegraphics[width=2\columnwidth]{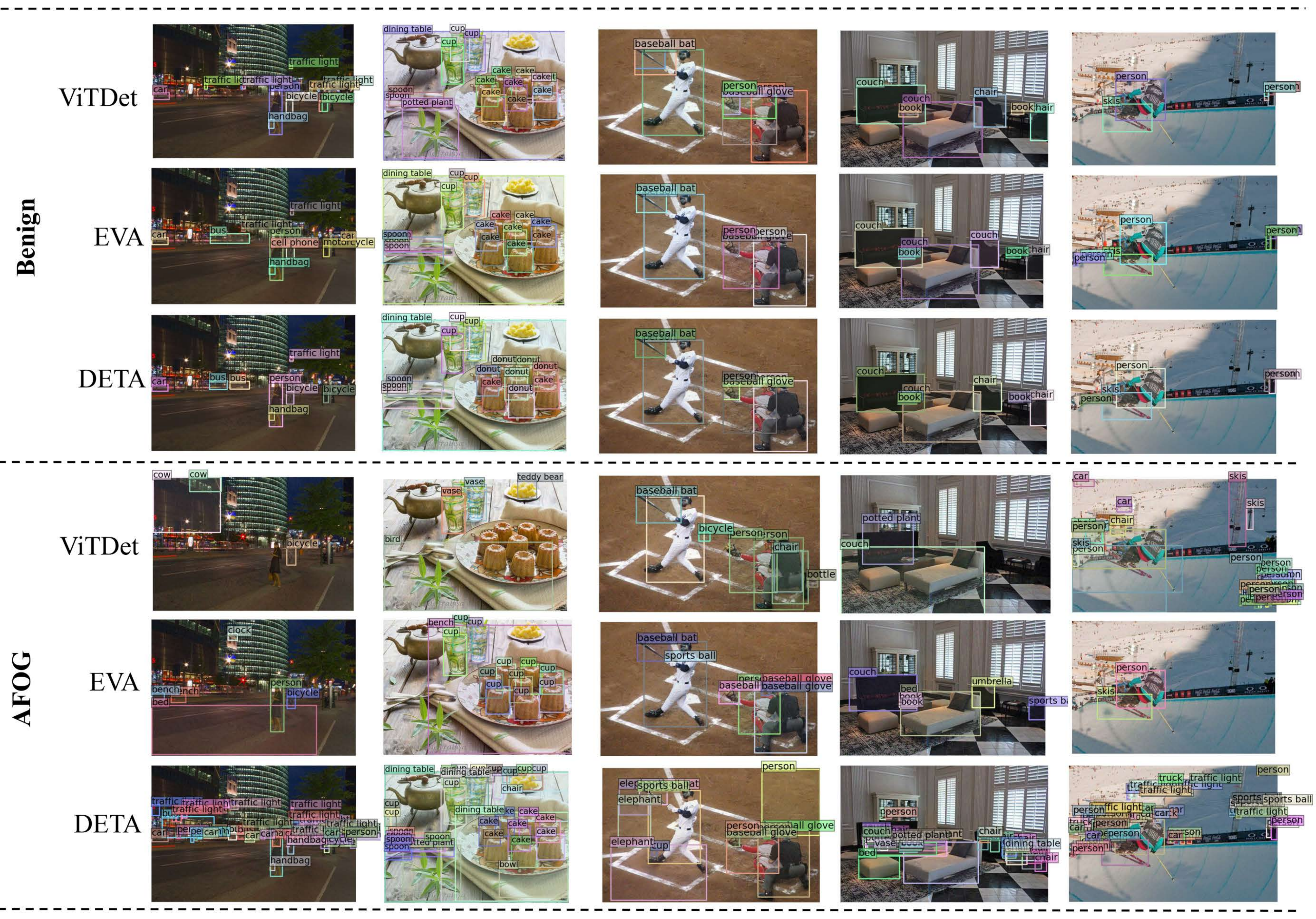}
    \label{fig:group-c-examples}
    \caption{Comparison of Benign and AFOG-disrupted predictions on five images with ViTDet, EVA, and DETA. Although AFOG shows strong attack performance against various other classes across the five images, it also struggles to disrupt the "Person" class in the third and the fifth examples.}
\end{figure*}

\begin{figure*}[!ht]
    \centering
    \includegraphics[width=2.1\columnwidth]{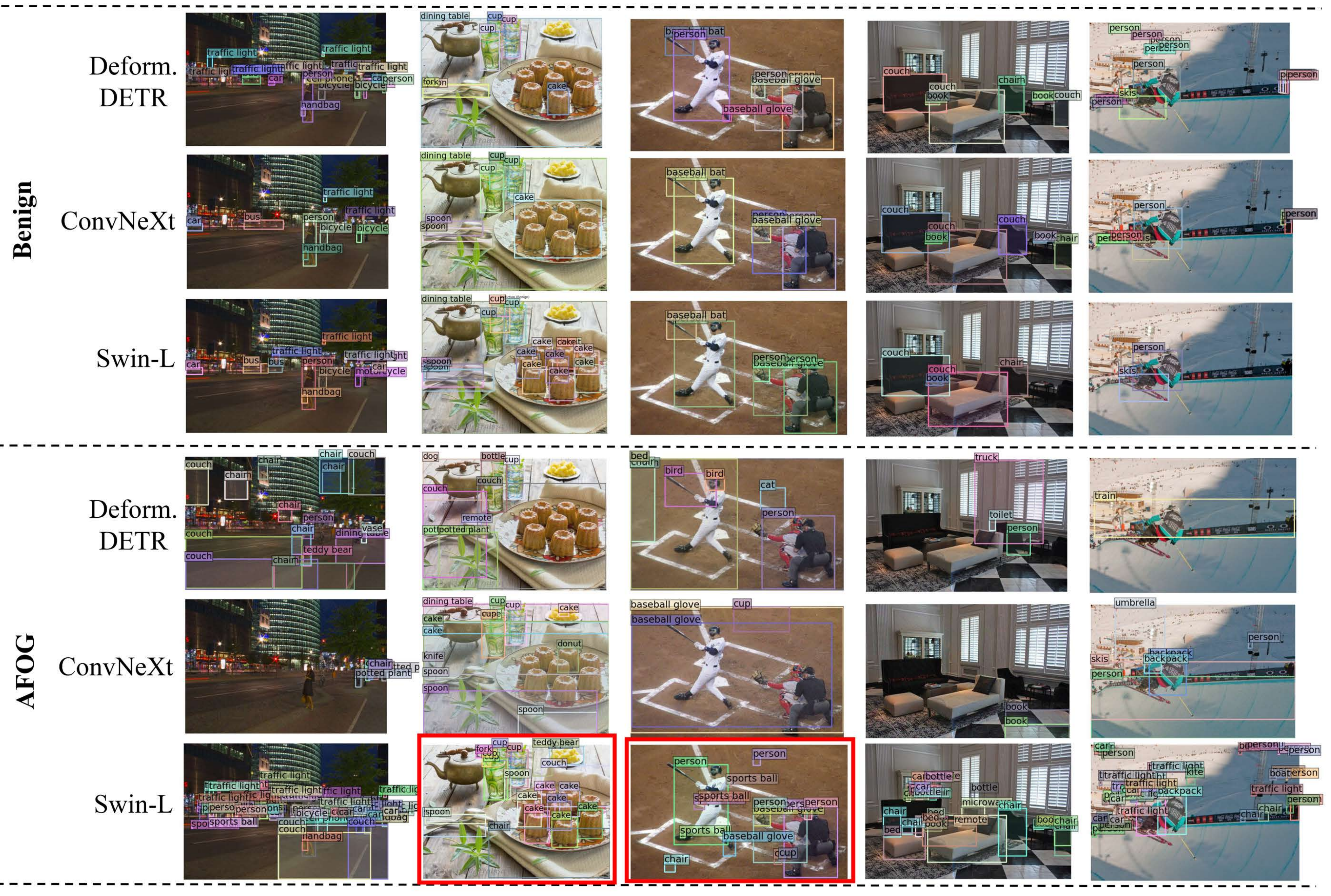}
    \caption{Comparison of Benign and AFOG-disrupted predictions on five example images with Deformable-DETR, ConvNeXt, and Swin-L. AFOG exhibits widely varying behavior for different models on the same image, such as against image three (Rows 4-6, Col. 3). We also indicate two instances where AFOG failed to disrupt Swin-L's performance, highlighted in red.}
    \label{fig:group-d-examples}
\end{figure*}

In Figures \ref{fig:group-a-examples}-\ref{fig:group-d-examples} we show visualizations of example object detection results under the benign scenario and AFOG attack for five COCO testdev images across all 12 transformers in our experiments. We partition these results into four groups to simplify comparison and highlight trends. Each group shows benign predictions for a set of detectors in the first rows, followed by AFOG-disrupted predictions in the next rows.

Figure \ref{fig:group-a-examples} compares detection results for FocalNet \cite{yang2022focal}, InternImage \cite{wang2022internimage}, DINO-ResNet50 \cite{he2016resnet}, and AlignDETR \cite{cai2023aligndetr}. We observe that, although these models each have unique architectures, they have nearly identical benign prediction behavior across five example images (Rows 1-4). We further observe that AFOG's learnable attention mechanism probes and exploits the individual weaknesses of each model, inducing unique disrupted behaviors in each (Rows 5-8). Although AFOG's induced malicious behavior can have similar themes, such as vanishing most detections in the first image (Col. 1) or causing a large ``Person" class detection in the third image (Col. 3), the details of this behavior change between models. We speculate that this is a result of AFOG's architecture-agnostic adaptability.

Figure \ref{fig:group-b-examples} compares detection results for DETR \cite{carion2020} with a ResNet-50 backbone and DETR with a ResNet-101 backbone. Rows 1-2 show the benign behavior of the two models. We observe that nearly identical architectures produce similar detections. However, AFOG causes spurious detections that differ between the two models (Rows 3-4). In image two (Col. 2) AFOG induces a large ``Couch" prediction in both models by disrupting both class and bounding box losses. Similarly, AFOG induces several small ``Cat" detections in image three (Col. 3) for both models. In these cases, AFOG has learned similar perturbations for similar architectures. Images 4 and 5 (Cols. 4-5) exhibit very different behavior under AFOG attack, demonstrating how AFOG's learnable attention is also able to exploit fine-grained differences in victim models as necessary.

Figure \textcolor{red}{12} compares detection results for VitDet \cite{li2022vitdet}, EVA \cite{EVA}, and DETA \cite{ouyangzhang2022nms}. We compare AFOG failure modes between these three models, and make two observations. (i) AFOG shows similar weaknesses against some objects across the three models, such as the ``Person" in image five (Col. 5) and the ``Baseball Bat" in image two (Col. 2). (ii) AFOG exhibits unique failure behavior for other objects, such as the ``Couch" object in image four (Col. 4). Against ViTDet (Row 4, Col. 4), AFOG fails to disrupt the "Couch" class label, attacking only the bounding box. Against EVA (Row 5, Col. 4), AFOG fails to disrupt the bounding box and changes ``Couch" to a similar ``Bed" label. We speculate that AFOG's ability to probe for weak points and adapt to different models may also lead to this kind of unpredictable failure behavior. 

Figure \ref{fig:group-d-examples} compares detection results for the remaining detectors from our experiments: Deformable-DETR \cite{zhu2020deformable}, ConvNeXt \cite{liu2022convnext}, and Swin-L \cite{liu2021Swin}. As with other models, we observe very different detection behavior under AFOG attacks between models. For example, we contrast Deformable-DETR's detections for image three (Row 4, Col. 3) with ConvNeXt's detections for the same image (Row 5, Col. 3), noting that the two have little in common. We also observe that AFOG largely fails to disrupt images two and three against Swin (Row 6, Cols. 2-3), leaving multiple class and bounding box detections intact. We highlight this image in red for emphasis.


\subsection{Comparison to Other Attacks}
\noindent
We next compare our AFOG attack with four existing representative victim-based detection attacks against FRCNN, a two-stage CNN-based object detector. 

\begin{figure*}[h]
    \centering
    \includegraphics[width=2.2\columnwidth]{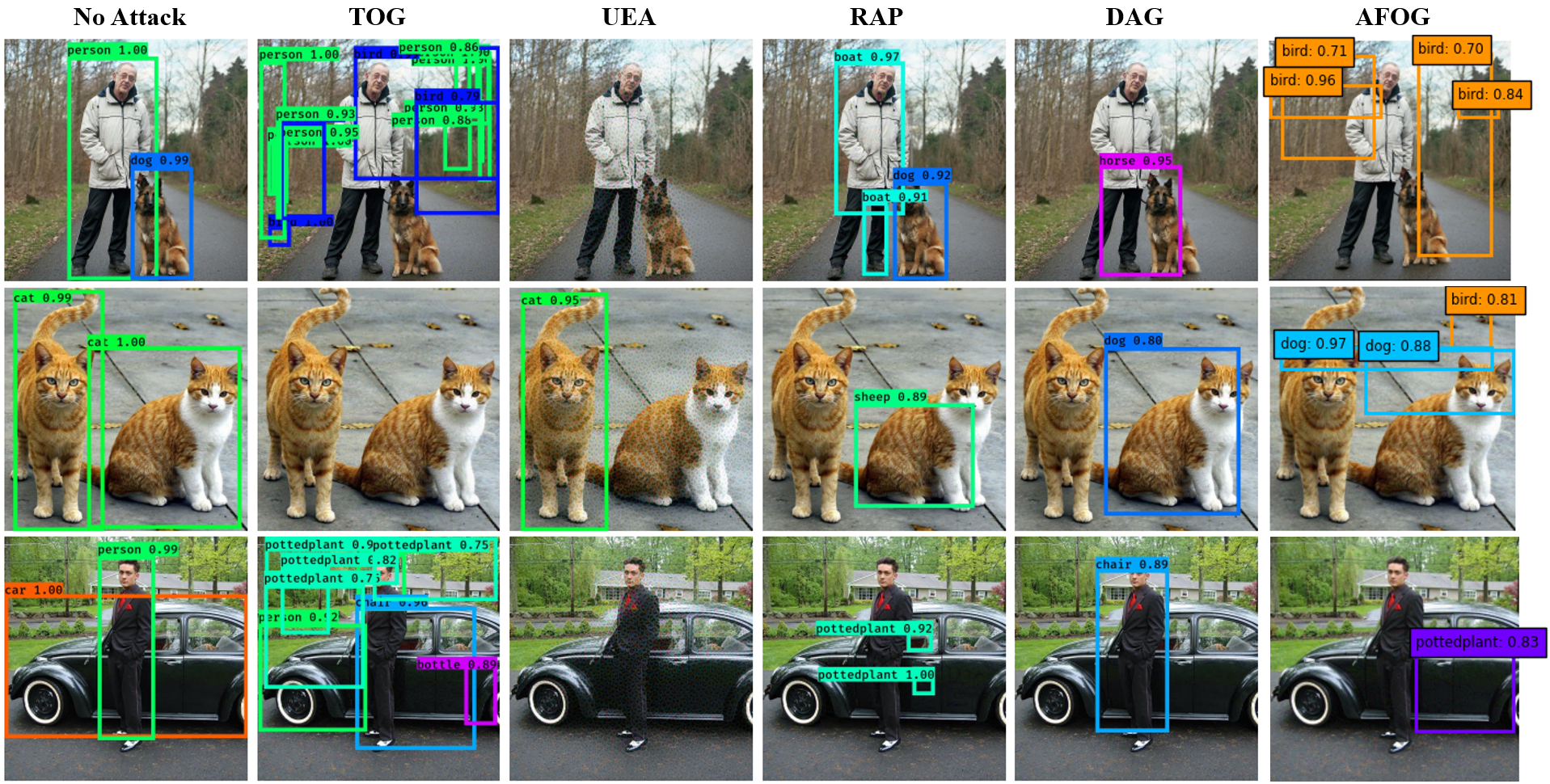}
    \caption{Comparison of three images across Benign, AFOG, TOG, UEA, RAP, and DAG cases. AFOG displays superior class and bounding-box disruption. Figure adapted from \cite{chow2020understanding}}
    \label{fig:reg-compare}
\end{figure*}

Figure \ref{fig:reg-compare} provides three examples from the Pascal VOC test set under the benign scenario (Column 1) and compares our AFOG attack (Column 6) with four well-known attacks: TOG \cite{chow2020tog} in Column 2, UEA \cite{wei2019uea} in Column 3, RAP \cite{li2018rap} in Column 4, and DAG \cite{xie2017dag} in Column 5. 
We make two observations. (i) All five attacks are successful in attacking the example images for which FRCNN can return correct detections under the benign scenario (Column 1). (ii) AFOG displays superior multi-box disruption, whereas other attacks produce misclassification errors with correct bounding boxes. 

\begin{figure*}[h]
    \centering
    \includegraphics[width=1.8\columnwidth]{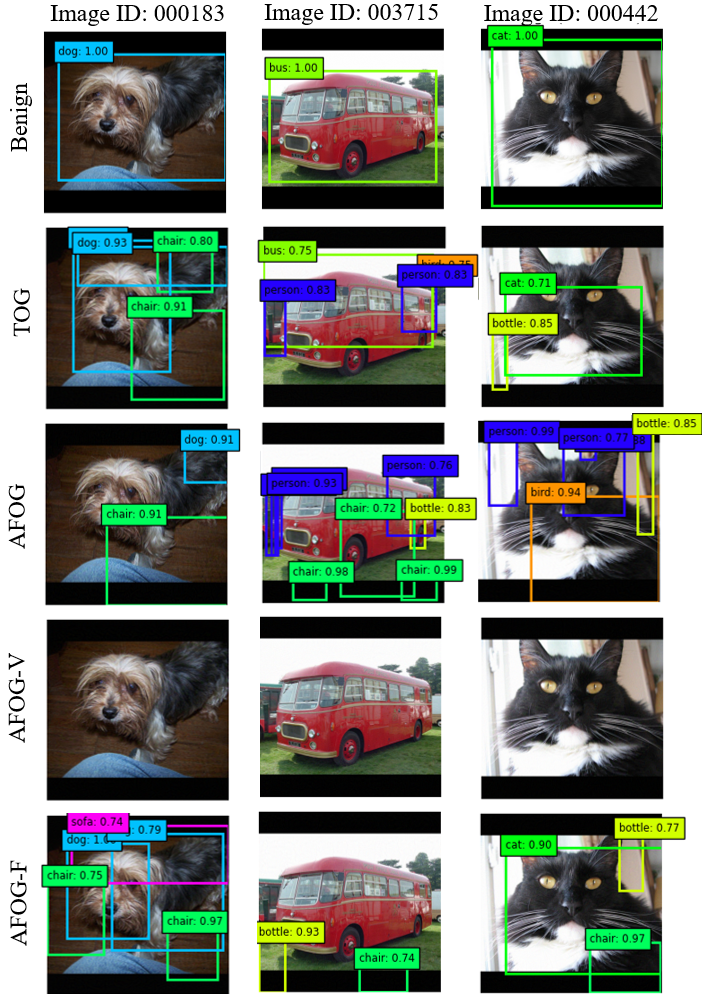}
    \caption{Three example images from the VOC dataset where AFOG outperforms TOG.}
    \label{fig:tog_comparison}
\end{figure*}

Figure \ref{fig:tog_comparison} further compares AFOG, AFOG-V and AFOG-F with TOG on three example images in the Pascal VOC test set. For all three examples, TOG (Row 2) fails to attack the FRCNN detector on the foreground objects (dog, bus, cat) and in contrast, AFOG (Row 3) is successful. AFOG-V (Row 4) shows the adverse effect of AFOG vanishing, and AFOG-F (Row 5) shows the adverse effect of AFOG fabrication. 

\end{document}